\documentclass[jair,twoside,11pt,theapa]{article}

\usepackage{jair, theapa, rawfonts}
\usepackage{multirow, graphicx}
\usepackage{caption, subcaption}
\usepackage{arydshln,url}
\usepackage{todonotes}
\usepackage{soul}
\usepackage{eucal}
\usepackage{amsmath,amssymb}
\usepackage{trackchanges}
\usepackage{footmisc}

\usepackage{color}

\DefineFNsymbols{mySymbols}{{\ensuremath\dagger}{\ensuremath\ddagger}\S\P
   *{**}{\ensuremath{\dagger\dagger}}{\ensuremath{\ddagger\ddagger}}}
\setfnsymbol{mySymbols}

\addeditor{Erik}
\soulregister\citep7

\jairheading{?}{2017}{?-??}{?/17}{?/17}
\ShortHeadings{Revisiting the ALE: Evaluation Protocols and Open Problems}
{Machado, Bellemare, Talvitie, Veness, Hausknecht, \& Bowling}
\firstpageno{1}

\def \cbar {\, | \,}
\def \cS {\mathcal{S}}
\def \cA {\mathcal{A}}
\def \cO {\mathcal{O}}
\def \bR {\mathbb{R}}

\newcommand{\citep}[1]{\cite{#1}}
\newcommand{\citet}[1]{\shortciteauthor{#1} \citeyear{#1}}
\newcommand{\gamename}[1]{\textsc{#1}}

\DeclareMathOperator*{\argmax}{arg\,max}

\begin{document}

\title{Revisiting the Arcade Learning Environment:\\Evaluation Protocols and Open Problems for General Agents}

\author{\name Marlos C. Machado \email machado@ualberta.ca \\
            \addr University of Alberta, Edmonton, Canada
            \AND
            \name Marc G. Bellemare \email bellemare@google.com \\
            \addr Google Brain, Montr\'eal, Canada\thanks{\ \ Work performed at DeepMind.}
            \AND
            \name Erik Talvitie \email erik.talvitie@fandm.edu \\
            \addr Franklin \& Marshall College, Lancaster, USA 
            \AND
            \name Joel Veness\email aixi@google.com \\
            \addr DeepMind, London, United Kingdom 
            \AND
            \name Matthew Hausknecht\email matthew.hausknecht@microsoft.com \\
            \addr Microsoft Research, Redmond, USA
            \AND
            \name Michael Bowling \email mbowling@ualberta.ca \\
            \addr University of Alberta, Edmonton, Canada\\
            \addr DeepMind, Edmonton, Canada}

\maketitle

\begin{abstract}
The Arcade Learning Environment (ALE) is an evaluation platform that poses the challenge of building AI agents with general competency across dozens of Atari 2600 games. It supports a variety of different problem settings and it has been receiving increasing attention from the scientific community, leading to some high-profile success stories such as the much publicized Deep Q-Networks (DQN). In this article we take a big picture look at how the ALE is being used by the research community. We show how diverse the evaluation methodologies in the ALE have become with time, and highlight some key concerns when evaluating agents in the ALE. We use this discussion to present some methodological best practices and provide new benchmark results using these best practices. To further the progress in the field, we introduce a new version of the ALE that supports multiple game modes and provides a form of stochasticity we call sticky actions. We conclude this big picture look by revisiting challenges posed when the ALE was introduced, summarizing the state-of-the-art in various problems and highlighting problems that remain open.
  \end{abstract}


\section{Introduction}
\label{Introduction}

The Arcade Learning Environment (ALE) is both a challenge problem and a platform for evaluating general competency in artificial intelligence (AI). Originally proposed by \citeauthor{Bellemare_JAIR13}~\citeyear{Bellemare_JAIR13}, the ALE makes available dozens of Atari 2600 games for agent evaluation. The agent is expected to do well in as many games as possible without game-specific information, generally perceiving the world through a video stream.
Atari 2600 games are excellent environments for evaluating AI agents for three main reasons: 1) they are varied enough to provide multiple different tasks, requiring general competence, 2) they are interesting and challenging for humans, and 3) they are free of experimenter's bias, having been developed by an independent party.

The usefulness of the ALE is reflected in the amount of attention it has received from the scientific community. The number of papers using the ALE as a testbed has exploded in recent years. This has resulted in some high-profile success stories, such as the much publicized Deep Q-Networks (DQN), the first algorithm to achieve human-level control in a large fraction of Atari 2600 games \shortcite{Mnih_Nature15}. This interest has also led to the first dedicated workshop on the topic, the AAAI Workshop on Learning for General Competency in Video Games \shortcite{Albrecht_AIM15}. Several of the ideas presented in this article were first discussed at this workshop, such as the need for standardizing evaluation and for distinguishing open-loop behaviours from closed-loop ones.

Given the ALE's increasing importance in the AI literature, this article aims to be a ``check-up'' for the Arcade Learning Environment, taking a big picture look at how the ALE is being used by researchers. The primary goal is to highlight some subtle issues that are often overlooked and propose some small course corrections to maximize the scientific value of future research based on this testbed. The ALE has incentivized the AI community to build more generally competent agents. The lessons learned from that experience may help further that progress and may inform best practices as other testbeds for general competency are developed (\emph{e.g.}, \shortciteauthor{levine13general}, 2013; \shortciteauthor{beattie16deepmindlab}, 2016; \shortciteauthor{brockman16openai}, 2016; \shortciteauthor{johnson16malmo}, 2016). 

The main contributions of this article are: 1)~To discuss the different evaluation methods present in the literature and to identify, for the typical reinforcement learning setting, some methodological best practices gleaned from experience with the ALE (Sections~\ref{sec:diff_goals}~and~\ref{sec:summarizing_performance}). 2)~To address concerns regarding the deterministic dynamics of previous versions of the platform, by introducing a new version of the ALE that supports a form of stochasticity we call \emph{sticky actions} (Section~\ref{sec:determinism_stochasticity}). 3)~To provide new benchmark results in the reinforcement learning setting that ease comparison and reproducibility of experiments in the ALE. These benchmark results also encourage the development of sample efficient algorithms (Section~\ref{sec:benchmark}). 4)~To revisit challenges posed when the ALE was introduced, summarizing the state-of-the-art in various problems and highlighting problems that are currently open (Section~\ref{sec:open_prob}). 5) To introduce a new feature to the platform that allows existent environments to be instantiated in multiple difficult levels and game modes (Section~\ref{sec:modes_diff_ale})

\section{Background}
\label{sec:background}

In this section we introduce the formalism behind reinforcement learning~\cite{Sutton98}, as well as how it is instantiated in the Arcade Learning Environment. We also present the two most common value function representations used in reinforcement learning for Atari 2600 games: linear approximation and neural networks. As a convention, we indicate scalar-valued random variables by capital letters (\emph{e.g.}, $S_t$, $R_t$), vectors by bold lowercase letters (\emph{e.g.}, $\boldsymbol{\theta}, \boldsymbol{\phi}$), functions by non-bold lowercase letters (\emph{e.g.}, $v$, $q$), and sets with a calligraphic font (\emph{e.g.}, $\mathcal{S}, \mathcal{A}$).

\subsection{Setting}
We consider an agent interacting with its environment in a sequential manner, aiming to maximize cumulative reward. It is often assumed that the environment satisfies the Markov property and is modeled as a Markov decision process (MDP). An MDP  is formally defined as a 4-tuple $\langle \cS, \cA, p, r\rangle$. Starting from state $S_0 \in \mathcal{S}$, at each step the agent takes an action $A_t \in \cA$, to which the environment responds with a state $S_t \in \cS$, according to a transition probability kernel $p(s' \cbar s, a) \doteq \Pr(S_{t+1} = s' \cbar S_t = s, A_t = a)$, and a reward $R_{t+1}$, which is generated by the function $r(s, a, s') \doteq r(S_{t} = s, A_t = a, S_{t+1} = s')\in \bR$.

In the context of the ALE, an action is the composition of a joystick direction and an optional button press. The agent observes a reward signal, which is typically the change in the player's score (the difference in score between the previous time step and the current time step), and an observation $O_t \in \cO$ of the environment. This observation can be a single $210 \times 160$ image and/or the current 1024-bit RAM state. Because a single image typically does not satisfy the Markov property, we distinguish between observations and the environment state, with the RAM data being the real state of the emulator.\footnote{The internal emulator state also includes registers and timers, but the RAM information and joystick inputs are sufficient to infer the next emulator state.} A frame (as a unit of time) corresponds to 1/60th of a second, the time interval between two consecutive images rendered to the television screen. The ALE is deterministic: given a particular emulator state $s$ and a joystick input $a$ there is a unique resulting next state $s'$, that is, $p(s' \cbar s, a) = 1$. We will return to this important characteristic in Section~\ref{sec:determinism_stochasticity}.

Agents interact with the ALE in an episodic fashion. An episode begins by resetting the ALE to its initial configuration, and ends at a natural endpoint of a game's playthrough (this often corresponds to the player losing their last life). The primary measure of an agent's performance is the score achieved during an episode, namely the undiscounted sum of rewards for that episode.
While this performance measure is quite natural, it is important to realize that score, in and of itself, is not necessarily an indicator of AI progress. In some games, agents can maximize their score by ``getting stuck'' in a loop of ``small'' rewards, ignoring what human players would consider to be the game's main goal. Nevertheless, score is currently the most common measure of agent performance so we focus on it here.

Beyond the minimal interface described above, almost all agents designed for the ALE implement some form of reward normalization. The magnitude of rewards can vary wildly across games; transforming the reward to fit into a roughly uniform scale makes it more feasible to find game-independent meta-parameter settings. For instance, some agents divide every reward by the magnitude of the first non-zero reward value encountered, implicitly assuming that the first non-zero reward is ``typical'' \citep{Bellemare_JAIR13}. Others account only for the sign of the reward, replacing each reward value with -1, 0, or 1, accordingly \citep{Mnih_Nature15}. Most agents also employ some form of hard-coded preprocessing to simplify the learning and acting process.  We briefly review the three most common preprocessing steps as they will play a role in the subsequent discussion. 1) \emph{Frame skipping}~\citep{naddaf10game} restricts the agent's decision points by repeating a selected action for $k$ consecutive frames. Frame skipping results in a simpler reinforcement learning problem and speeds up execution; values of $k=4$ and $k=5$ have been commonly used in the literature. 2) \emph{Color averaging} \citep{Bellemare_JAIR13} and \emph{frame pooling} \citep{Mnih_Nature15} are two image-based mechanisms to flatten two successive frames into a single one in order to reduce visual artifacts resulting from limitations of the Atari 2600 hardware -- by leveraging the slow decay property of phosphors on 1970s televisions, objects on the screen could be displayed every other frame without compromising the game's visual aspect \citep{montfort09racing}. 
Effectively, color averaging and frame pooling remove the most benign form of partial observability in the ALE. Finally, 3) \emph{frame stacking} \citep{Mnih_Nature15} concatenates previous frames with the most recent in order to construct a richer observation space for the agent. Frame stacking also reduces the degree of partial observability in the ALE, making it possible for the agent to detect the direction of motion in objects.

\subsection{Control in the Arcade Learning Environment}

The typical goal of reinforcement learning (RL) algorithms is to learn a \emph{policy} $\pi : \cS \times \cA \rightarrow[0, 1]$ that maps each state to a probability distribution over actions. Ideally, following the learned policy will maximize the discounted cumulative sum of rewards.\footnote{We use the \emph{discounted} sum of rewards in our formalism because this is commonly employed by agents in the ALE. Empirical evidence has shown that agents generally perform better when maximizing the \emph{discounted} cumulative sum of rewards, even though they are actually evaluated in the \emph{undiscounted} case. This formulation disincentivizes agents to postpone scoring.} Many RL algorithms accomplish this by learning an \emph{action-value} function $q_\pi : \cS \times \cA \rightarrow \bR$, which encodes the long-range value of taking action $a$ in state $s$ and then following policy $\pi$ thereafter. More specifically, $q_\pi(s, a) \doteq \mathbb{E}\left[\sum_{i=1}^\infty \gamma^{i-1}R_{t+i} \mid S_t = s, A_t = a\right]$, the expected discounted sum of rewards for some discount factor $\gamma \in [0, 1]$, where the expectation is over both the policy $\pi$ and the probability kernel $p$. However, in the ALE it is not feasible to learn an individual value for each state-action pair due to the large number of possible states. A common way to address this issue is to approximate the action-value function by parameterizing it with a set of weights $\boldsymbol{\theta} \in \bR^n$ such that $q_\pi(s,a) \approx q_\pi(s, a, \boldsymbol{\theta})$. We discuss below two approaches to value function approximation that have been successfully applied to the games available in the ALE. We focus on these particular methods because they are by now well-established, well-understood, achieve a reasonable level of performance, and reflect the issues we study here.

The first approach is to design a function that, given an observation, outputs a vector $\boldsymbol{\phi}(s, a)$ denoting a feature representation of the state $s$ when taking action $a$. With this approach, we estimate $q_\pi$ through a linear function approximator $q_\pi(s, a, \boldsymbol{\theta}) = \boldsymbol{\theta}^\top \boldsymbol{\phi}(s,a)$. Sarsa($\lambda$)~\cite{Rummery94} is a control algorithm that learns an approximate action-value function of a continually improving policy $\pi$. As states are visited, and rewards are observed, $q_\pi$ is updated and $\pi$ is consequently improved. The update equations are:
\begin{eqnarray*}
\delta_t     & = & R_{t+1} + \gamma \boldsymbol{\theta}_t^\top \boldsymbol{\phi}(s_{t + 1}, a_{t+1}) - \boldsymbol{\theta}_t^\top \boldsymbol{\phi}(s_{t}, a_{t})\\
\boldsymbol{e}_t          & = & \gamma \lambda \boldsymbol{e}_{t-1} + \boldsymbol{\phi}(s_{t}, a_{t})\\
\boldsymbol{\theta}_{t+1} & = & \boldsymbol{\theta}_t + \alpha \delta_t \boldsymbol{e}_t
\end{eqnarray*}
where $\alpha$ denotes the step-size, $\boldsymbol{e}_t$ the eligibility trace vector ($\boldsymbol{e}_{-1} \doteq \boldsymbol{0}$), $\delta_t$ the temporal difference error, and $\gamma$ the discount factor. The first benchmarks in the ALE applied this approach with a variety of simple feature representations \citep{naddaf10game,Bellemare_NIPS12,Bellemare_JAIR13}. Recently, \citeauthor{Liang_AAMAS16}~\citeyear{Liang_AAMAS16} introduced a feature representation (Blob-PROST) that allows Sarsa($\lambda$) to achieve comparable performance to DQN (described below) in several Atari~2600 games. We refer to such an approach as Sarsa($\lambda$) $+$ Blob-PROST. Recently, \citet{Martin_IJCAI17} combined Sarsa($\lambda$) and the Blob-PROST features with a method for incentivizing exploration in hard games.

A recent trend in reinforcement learning is to use neural networks to estimate $q_\pi(s, a, \boldsymbol{\theta})$, substituting the requirement of a good handcrafted feature representation with the requirement of an effective network architecture and algorithm. \citeauthor{Mnih_Nature15}~\citeyear{Mnih_Nature15} introduced Deep Q-Networks (DQN), an algorithm that learns representations in a neural network composed of three hidden convolutional layers followed by a fully-connected hidden layer. The network weights are updated through backpropagation with the following update rule:
\begin{eqnarray*}
\boldsymbol{\theta}_{t+1} & = & \boldsymbol{\theta}_t + \alpha \Big[ R_{t+1} + \gamma \max_{a \in \cA} \tilde{q}(S_{t+1}, a, \boldsymbol{\theta}_t) - q(S_t, A_t, \boldsymbol{\theta}_t) \Big] \nabla_{\boldsymbol{\theta}_t} q(S_t, A_t, \boldsymbol{\theta}_t)
\end{eqnarray*}
where $\tilde{q}$ denotes the action-values estimated by a second network. This second network is updated less frequently for stability purposes. Additional components of the algorithm include clipping the rewards (as described above) and the use of experience replay \cite{Lin93} to decorrelate observations. DQN has inspired much follow-up work combining reinforcement learning and deep neural networks (\emph{e.g.}, \shortciteauthor{jaderberg17reinforcement}, 2017; \shortciteauthor{Mnih_ICML16}, 2016; \shortciteauthor{Schaul_ICLR16}, 2016; \shortciteauthor{Hasselt_AAAI16}, 2016).

\section{Divergent Evaluation Methodologies in the ALE}\label{sec:diff_goals}

The ALE has received significant attention since it was introduced as a platform to evaluate general competency in AI. Hundreds of papers have used the ALE as a testbed, employing many distinct experimental protocols for evaluating agents. Unfortunately, these different evaluation protocols are often not carefully distinguished, making direct comparisons difficult or misleading. In this section we discuss a number of methodological differences that have emerged in the literature. In subsequent sections we give special focus to two particularly important methodological issues: 1)~\emph{different metrics for summarizing agent performance}, and 2)~\emph{different mechanisms for injecting stochasticity in the environment}. 

The discussion about the divergence of evaluation protocols and the need for standardizing them first took place at the AAAI Workshop on Learning for General Competency in Video Games. One of the reasons that authors compare results generated with differing experimental protocols is the high computational cost of evaluating algorithms in the ALE -- it is difficult to re-evaluate existing approaches to ensure matching methodologies. For that reason it is perhaps especially important to establish a standard methodology for the ALE in order to reduce the cost of principled comparison and analysis. One of the main goals of this article is to propose such a standard, and to introduce benchmark results obtained under it for straightforward comparison to future work. 

\subsection{Methodological Differences}

To illustrate the diversity in evaluation protocols, we discuss some methodological differences found in the literature. While these differences may be individually benign, they are frequently ignored when comparing results, which undermines the validity of direct comparisons.

{\bf Episode termination.} In the initial ALE benchmark results \cite{Bellemare_JAIR13}, episodes terminate when the game is over. However, in some games the player has a number of ``lives'' which are lost one at a time. Terminating only when the game is over often makes it difficult for agents to learn the significance of losing a life. \citet{Mnih_Nature15} terminated training episodes when the agent lost a life, rather than when the game is over (evaluation episodes still lasted for the entire game). While this approach has the potential to teach an agent to avoid ``death,'' \citet{Bellemare_NIPS16} noted that it can in fact be detrimental to an agent's performance. Currently, both approaches are still common in the literature. We often see episodes terminating when the game is over (\emph{e.g.}, \shortciteauthor{Hausknecht_TCIAIG14}, 2014; \shortciteauthor{Liang_AAMAS16}, 2016; \shortciteauthor{Lipovetzky_IJCAI15}, 2015; \shortciteauthor{Martin_IJCAI17}, 2017), as well as when the agent loses a life (\emph{e.g.}, \shortciteauthor{Nair_Workshop15}, 2015; \shortciteauthor{Schaul_ICLR16}~2016; \shortciteauthor{Hasselt_AAAI16}, 2016). Considering the ideal of minimizing the use of game-specific information and the questionable utility of termination using the ``lives'' signal, \emph{we recommend that only the game over signal be used for termination}.

{\bf Setting of hyperparameters.} One of the primary goals of the ALE is to enable the evaluation of agents' general ability to learn in complex, high-dimensional decision-making problems. Ideally agents would be evaluated in entirely novel problems to test their generality, but this is of course impractical. With only 60 available games in the standard suite there is a risk that methods could ``overfit'' to the finite set of problems. In analogy to typical methodology in supervised learning, \citet{Bellemare_JAIR13} split games into ``training'' and ``test'' sets, only using results from training games for the purpose of selecting hyperparameters, then fully evaluating the agent in the test games only once hyperparameters have been selected. This methodology has been inconsistently applied in subsequent work -- for example, hyperparameters are sometimes selected using the entire suite of games, and in some cases hyperparameters are optimized on a per-game basis (\emph{e.g.}, \shortciteauthor{jaderberg17reinforcement}, 2017). For the sake of evaluating generality, \emph{we advocate for a train/test game split as a way to evaluate agents in problems they were not specifically tuned for}.

{\bf Measuring training data.} The first benchmarks in the ALE \cite{Bellemare_JAIR13} trained agents for a fixed number of episodes before evaluating them. This can be misleading since episode lengths differ from game to game. Worse yet, in many games the better an agent performs the longer episodes last. Thus, under this methodology, agents that learn a good policy early receive more training data overall than those that learn more slowly, potentially magnifying their differences. Recently it has become more common to measure the amount of training data in terms of the total number of frames experienced by the agent~\cite{Mnih_Nature15}, which aids reproducibility, inter-game analysis, and fair comparisons. That said, since performance is measured on a per-episode basis, it may not be advisable to end training in the middle of an episode. For example, \citet{Mnih_Nature15} interrupt the training as soon as the maximum number of frames is reached, while \citet{Liang_AAMAS16} pick a total number of training frames, and then train each agent until the end of the episode in which the total is exceeded. This typically results in a negligible number of extra frames of experience beyond the limit. Another important aspect to be taken into consideration is frame skipping, which is a common practice in the ALE but is not reported consistently in the literature. \emph{We advocate evaluating from full training episodes from a fixed number of frames}, as was done by \citet{Liang_AAMAS16}, and \emph{we advocate taking the number of skipped frames into consideration when measuring training data}, as the time scale in which the agent operates is also an algorithmic choice.

{\bf Summarizing learning performance.} When evaluating an agent in 60 games, it becomes necessary to compactly summarize the agent's performance in each game in order to make the results accessible and to facilitate comparisons. Authors have employed various statistics for summarizing agent performance and this diversity makes it difficult to directly compare reported results. \emph{We recommend reporting training performance at different intervals during learning.} We discuss this issue in more detail in Section \ref{sec:summarizing_performance}.

{\bf Injecting stochasticity.} The original Atari 2600 console had no source of entropy for generating pseudo-random numbers. The Arcade Learning Environment is also fully deterministic -- each game starts in the same state and outcomes are fully determined by the state and the action. As such, it is possible to achieve high scores by learning an open-loop policy, \emph{i.e.}, by simply memorizing a good action sequence, rather than learning to make good decisions in a variety of game scenarios \citep{Bellemare_IJCAI15}. Various approaches have been developed to add forms of stochasticity to the ALE dynamics in order to encourage and evaluate robustness in agents (\emph{e.g.}, \shortciteauthor{brockman16openai}, 2016; \shortciteauthor{Hausknecht_LGCVG15}, 2015; \shortciteauthor{Mnih_Nature15}, 2015; \shortciteauthor{Nair_Workshop15}, 2015). \emph{Our recommendation is to use \emph{sticky actions}, implemented in the latest version of the ALE}. We discuss this issue in more detail in Section \ref{sec:determinism_stochasticity}.

\section{Summarizing Learning Performance}~\label{sec:summarizing_performance}

One traditional goal in reinforcement learning is for agents to continually improve their performance as they obtain more data \shortcite{wilson85knowledge,thrun93lifelong,ring97child,singh04intrinsically,hutter05universal,sutton11horde}. Measuring the extent to which this is the case for a given agent can be a challenge, and this challenge is exacerbated in the Arcade Learning Environment, where the agent is evaluated across 60 games. When evaluating an agent in only a few problems, it is common practice to plot learning curves, which provide a rich description of the agent's performance: how quickly it learns, the highest performance it attains, the stability of its solutions, whether it is likely to continue to improve with more data, etc. 

While some have reported results in the ALE using learning curves (\emph{e.g.}, \shortciteauthor{Mnih_ICML16}~2016; \shortciteauthor{Ostrovski17}~2017; \shortciteauthor{Schaul_ICLR16}~2016), it is difficult to even effectively display, let alone comprehend and compare, 60 learning curves. For the sake of comparison and compact reporting, most researchers have applied various approaches to numerically summarize an agent's performance in each game (\emph{e.g.}, \shortciteauthor{Bellemare_JAIR13}, 2013; \shortciteauthor{Hausknecht_TCIAIG14}, 2014; \shortciteauthor{Munos_NIPS16}, 2016; \shortciteauthor{Nair_Workshop15}, 2015). Unfortunately, the variety of different summary statistics in results tables makes direct comparison difficult. In this section we consider some common performance measures seen in the literature and ultimately identify one as being particularly in line with the continual learning goal and advocate for it as the standard for reporting learning results in the ALE.

\subsection{Common Performance Measures}

Here we discuss some common summary statistics of learning performance that have been employed in the Arcade Learning Environment in the past.

\textbf{Evaluation after learning.} In the first ALE benchmark results, \citet{Bellemare_JAIR13} trained agents for a fixed training period, then evaluated the learned policy using the average score in a number of evaluation episodes with no learning. Naturally, a number of subsequent studies used this evaluation protocol (\emph{e.g.}, \shortciteauthor{Defazio_ARXIV13}, 2013; \shortciteauthor{Liang_AAMAS16}, 2016; \shortciteauthor{Martin_IJCAI17}, 2017). One downside to this approach is that it hides issues of sample efficiency, since agents are not evaluated during the entire training period. Furthermore, an agent can receive a high score using this metric without continually improving its performance. For instance, an agent could spend its training period in a purely exploratory mode, gathering information but performing poorly, and then at evaluation time switch to an exploitative mode. While the problem of developing a good policy during an unevaluated training period is an interesting one, in reinforcement learning the agent is typically expected to continually improve with experience. Importantly, $\epsilon$-greedy policies tend to perform better than greedy policies in the ALE \cite{Bellemare_JAIR13,Mnih_Nature15}. Therefore, this protocol does not necessarily benefit from turning off exploration during evaluation. In fact, often the reported results under this protocol do use $\epsilon$-greedy policies during evaluation.

\begin{figure}
  \centering
  \begin{subfigure}[b]{0.4\textwidth} 
    \includegraphics[width=\textwidth]{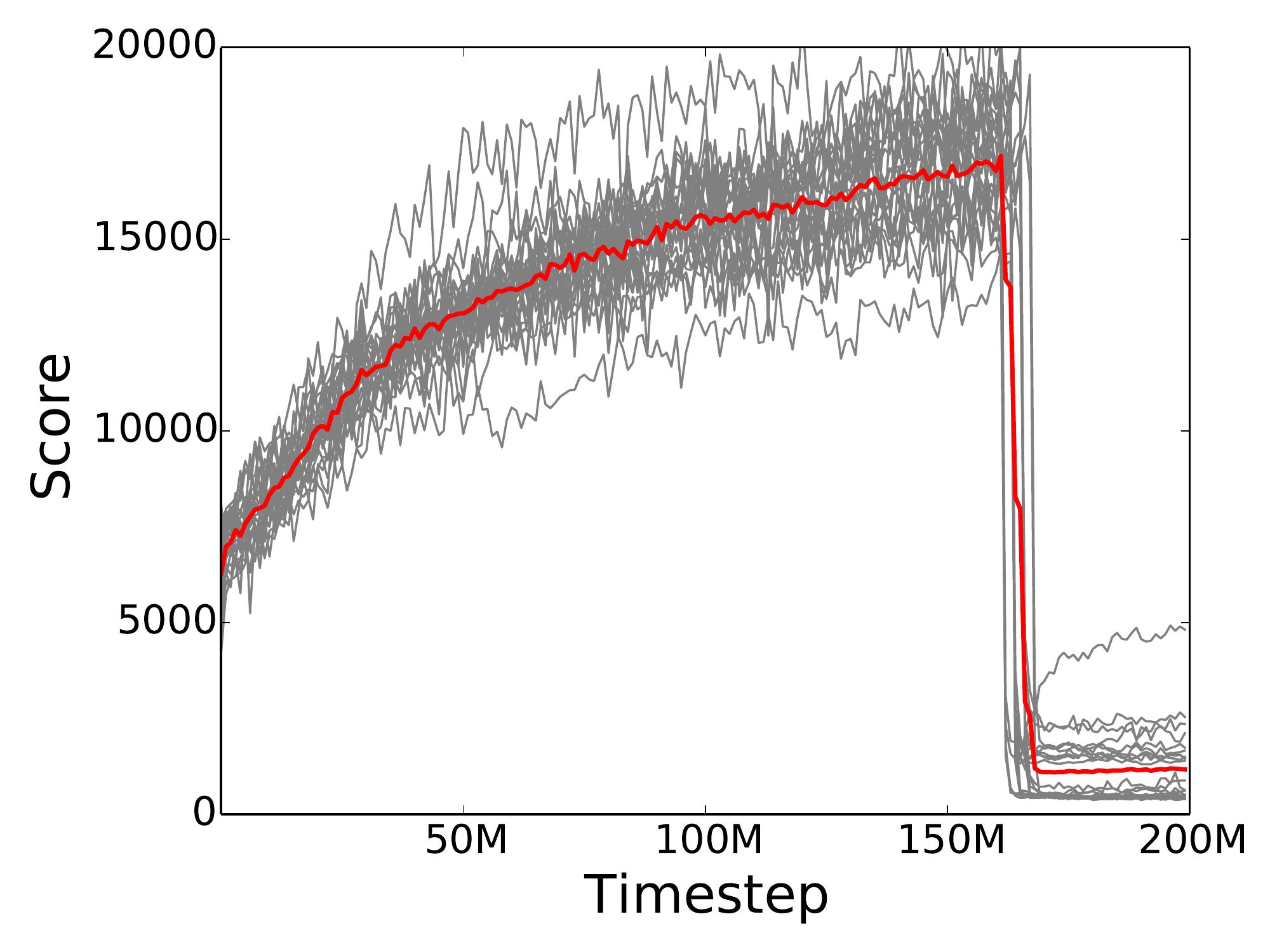} 
    \centering
    \caption{Sarsa($\lambda$) $+$ Blob-PROST} 
    \label{fig:centipedeSarsa} 
  \end{subfigure} ~ 
  \begin{subfigure}[b]{0.4\textwidth} 
    \includegraphics[width=\textwidth]{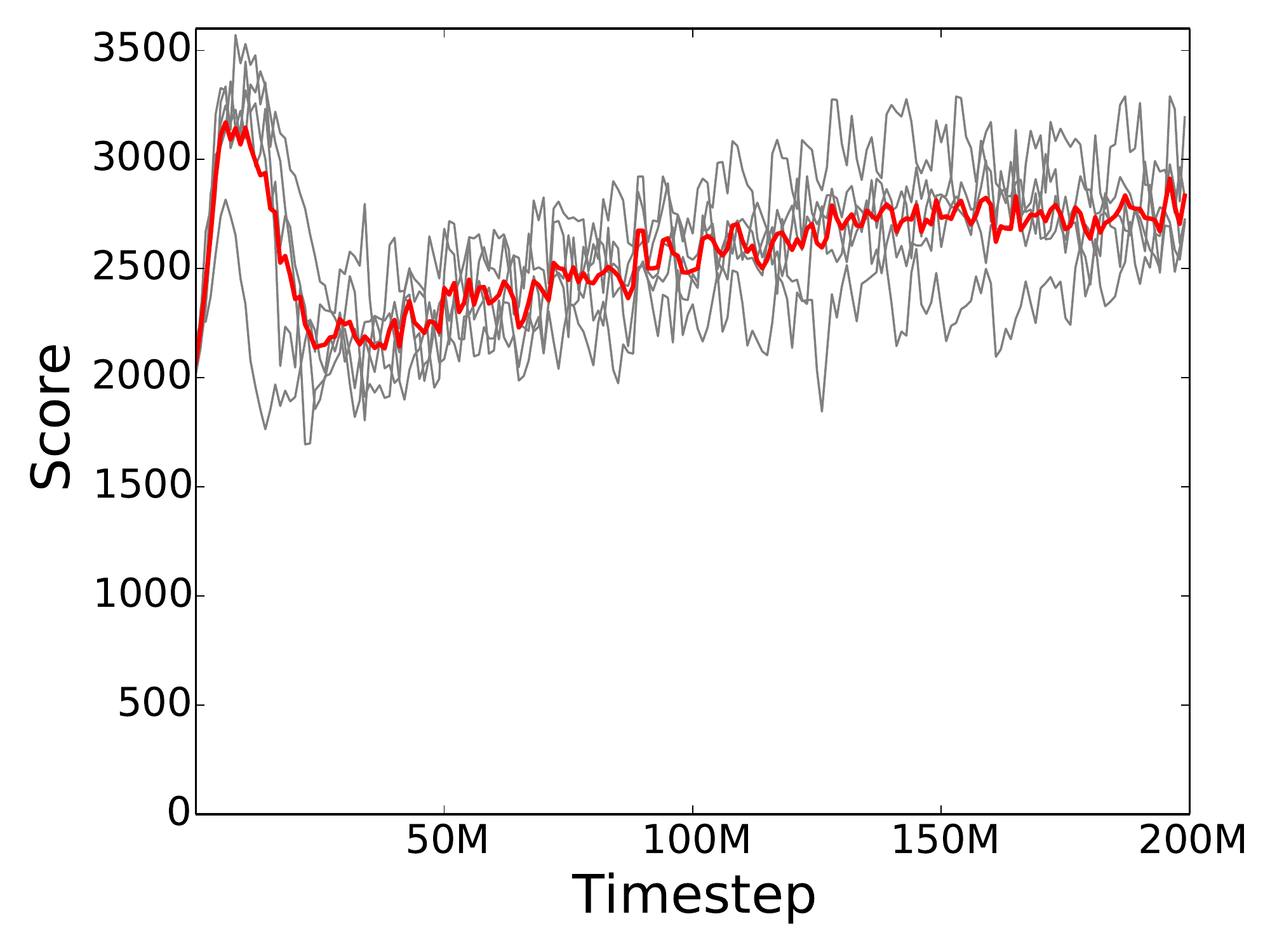}
    \centering
    \caption{DQN}
    \label{fig:centipedeDQN}
  \end{subfigure}
  \caption{Comparison between learning curves of DQN and Sarsa($\lambda$) $+$ Blob-PROST in \textsc{Centipede}. Notice the y-axes are not on the same scale. Each point corresponds to the average performance over the last one million episodes. Grey curves depict individual trials. The red curve depicts the average over all trials.}
  \label{fig:comparisonCentipede}
\end{figure}

\textbf{Evaluation of the best policy.} When evaluating Deep Q-Networks, \citet{Mnih_Nature15} also trained agents for a fixed training period. Along the way, they regularly evaluated the performance of the learned policy. At the end of the training period they evaluated the {\em best} policy in a number of evaluation episodes with no learning. A great deal of follow-up work has replicated this methodology (\emph{e.g.}, \shortciteauthor{Schaul_ICLR16}, 2016; \shortciteauthor{Hasselt_AAAI16}, 2016). This protocol retains the downsides of evaluation after learning, and adds an additional one: it does not evaluate the {\em stability} of the agent's learning progress. Figure~\ref{fig:comparisonCentipede} illustrates the importance of this issue by showing different learning curves in the game \textsc{Centipede}. On one hand, Sarsa$(\lambda)$ + Blob-PROST achieves a high score early on but then becomes unstable and fails to retain this successful policy. DQN's best score is much lower but it is also more stable (though not perfectly so). Reporting the performance of the best policy fails to recognize the plummeting behavior of both algorithms and DQN's more stable performance. Note also that the best score achieved across training is a statistically biased estimate of an agent's best performance: to avoid this bias, one should perform a second, independent evaluation of the agent at that particular point in time, as reported by \citet{wang16dueling}.

\textbf{Area under the learning curve.} Recently, eschewing an explicit evaluation phase, \citeauthor{Stadie_ARXIV15}~\citeyear{Stadie_ARXIV15} proposed the area under the learning curve as an evaluation metric. Intuitively, the area under the learning curve is generally proportional to how long a method achieves ``good'' performance, \emph{i.e.}, the average performance during training. Methods that only have performance spikes and methods that are unstable generally perform poorly under such metric. However, area under the learning curve does not capture the ``plummeting'' behavior illustrated in Figure~\ref{fig:comparisonCentipede}. For example, in this case, Sarsa($\lambda$) + Blob-PROST looks much better than DQN using this metric. Furthermore, area under the curve cannot distinguish a high-variance, unstable learning process from steady progress towards a good policy, even though we typically prefer the latter. 

\subsection{Proposal: Performance During Training}

The performance metric we propose as a standard is simple and has been adopted before (e.g. \shortciteauthor{Bellemare_AAAI12}~2012). At the end of training (and ideally at other points as well) report the average performance of the last $k$ episodes. This protocol does not use the explicit evaluation phase, thus requiring an agent to perform well while it is learning. This better aligns the performance metric with the goal of continual learning while also simplifying experimental methodology. Unstable methods that exhibit spiking and/or plummeting learning curves will score poorly compared to those that stably and continually improve, even if they perform well during most of training. 

Another advantage is that this metric is well-suited for analysis of an algorithm's sample efficiency. While the agent's performance near the end of training is typically of most interest, it is also straightforward to report the same statistic at various points during training, effectively summarizing the learning curve with a few selected points along the curve. Furthermore, if researchers make their full learning curve data publicly available, others can easily perform post-hoc analysis for the sake of comparison for any amount of training without having to fully re-evaluate existing methods. Currently, it is fairly standard to train agents for 200 million frames, in order to facilitate comparison with the DQN results reported by \citet{Mnih_Nature15}. This is equivalent to approximately 38 days of real-time gameplay and even at fast frame rates represents a significant computational expense. By reporting performance at multiple points during training, researchers can easily draw comparisons earlier in the learning process, reducing the computational burden of evaluating agents.

In accordance with this proposal, the benchmark results we present in Section~\ref{sec:benchmark} report the agent's average score of the last 100 episodes before the agent reaches 10, 50, 100, and 200 million frames and our full learning curve data is publicly available\footnote{\url{http://www.marcgbellemare.info/static/data/machado17revisiting.zip}}. This allows us to derive insight regarding the learning rate and stability of the algorithms and will offer flexibility to researchers wishing to compare to these benchmarks in the future.

\section{Determinism and Stochasticity in the Arcade Learning Environment}~\label{sec:determinism_stochasticity}

\begin{figure}[h]
  \centering
  \begin{subfigure}[b]{0.32\textwidth} 
    \raisebox{2.75mm}{\includegraphics[width=\textwidth]{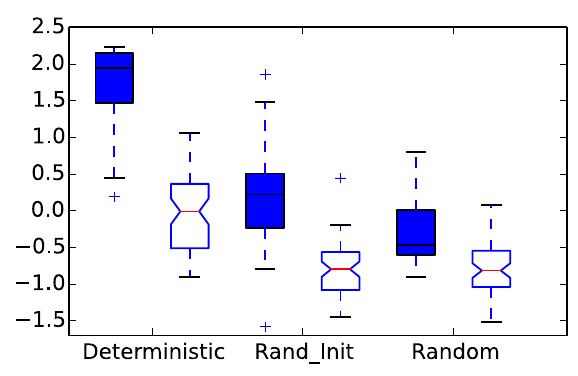}}
    \centering
  \end{subfigure} ~ 
  \begin{subfigure}[b]{0.31\textwidth} 
    \includegraphics[width=\textwidth]{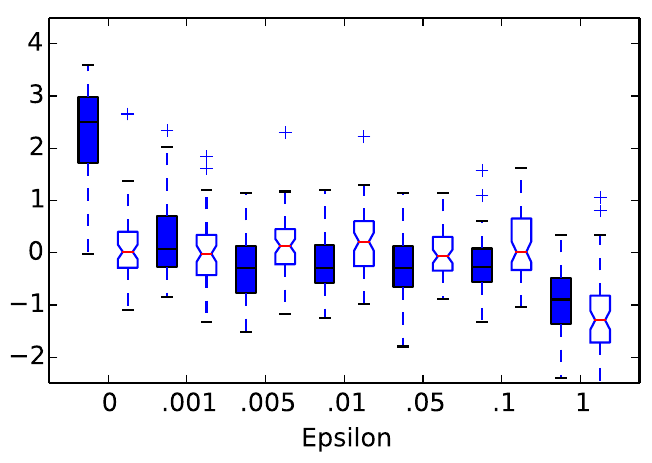}
    \centering
  \end{subfigure}
  \begin{subfigure}[b]{0.31\textwidth} 
    \includegraphics[width=\textwidth]{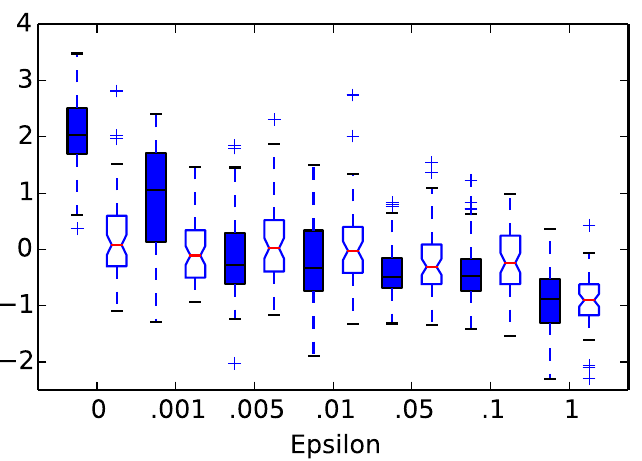}
    \centering
  \end{subfigure}
  \caption{Final performance of a HyperNEAT agent under various models of stochasticity in ALE. Each plot corresponds to a different stochasticity model. Retangular boxplots correspond to memorizing-NEAT while hollow, pinched boxplots correspond to randomized-NEAT~\cite{Hausknecht_TCIAIG14}. Each boxplot represents a single evaluation of 61 Atari 2600 games. Z-Score normalization is applied to normalize the per-game scores. The agent's overall performance is depicted in the $y$-axis while the amount of stochasticity in the environment increases along the $x$-axis. The first figure depicts the impact of random no-ops at the beginning of the game. Reference scores for a fully deterministic and fully random environments are provided. The second graph depicts the performance of both algorithms when forced to select actions $\epsilon$-greedily for different values of $\epsilon$. The third graph depicts the performance of both algorithms when forced to repeat the previous action with probability $\epsilon$ (equivalent to sticky actions). Reproduced from \citet{Hausknecht_LGCVG15}.}
  \label{fig:hausknecht}
\end{figure}

In almost all games, the dynamics within Stella itself (the Atari 2600 VCS emulator embedded within the ALE) are deterministic given the agent's actions. The agent always starts at the same initial state, and a given sequence of actions always leads to the same outcome. \citet{Bellemare_IJCAI15} and \citet{Braylan_LGCVG15} showed that this determinism can be exploited by agents that simply memorize an effective sequence of actions, attaining state-of-the-art scores while ignoring the agent's perceived state altogether. Such an approach is not likely to be successful beyond the ALE -- in most problems of interest it is difficult, if not impossible, to exactly reproduce a specific state-action sequence, and closed-loop decision-making is required. An agent that relies upon the determinism of the ALE may achieve high scores, but may also be highly sensitive to small perturbations. For example, \citet{Hausknecht_LGCVG15} analyzed the role of determinism in the success of HyperNEAT-GGP~\shortcite{Hausknecht_TCIAIG14}. Figure~\ref{fig:hausknecht} shows that {\em memorizing-NEAT} (solid boxes) performs significantly worse under multiple forms of mild stochasticity, whereas {\em randomized-NEAT} (hollow, pinched boxes), which is trained with some stochastic perturbations, performs worse in the deterministic setting, but is more robust to various forms of stochasticity. As an evaluation platform, the deterministic ALE does not effectively distinguish between agents that learn robust, closed-loop policies from brittle memorization-based agents.

Recognizing this limitation in earlier versions of the ALE, many researchers have augmented the standard behavior of the ALE to evaluate the robustness of their agents and to discourage memorization (\emph{e.g.}, injecting stochasticity, \shortciteauthor{Hausknecht_LGCVG15}, 2015; no-ops, \shortciteauthor{Mnih_Nature15}, 2015; human starts, \shortciteauthor{Nair_Workshop15}, 2015; random frame skips, \shortciteauthor{brockman16openai}, 2016). Again, this wide range of experimental protocols makes direct comparison of results difficult. We believe the research community would benefit from a single standard protocol that empirically distinguishes between brittle, open-loop solutions and robust, closed-loop solutions.

In this section we discuss the Brute~(first briefly introduced in \shortciteauthor{Bellemare_IJCAI15}, 2015) as an example of an algorithm that explicitly and effectively exploits the environment's determinism. We present results in five Atari 2600 games comparing the Brute's performance with traditionally successful reinforcement learning methods. We then introduce the {\em sticky actions} method for injecting stochasticity into the ALE and show that it effectively distinguishes the Brute from methods that learn more robust policies. We also discuss pros and cons of several alternative experimental protocols aimed at discouraging open-loop policies, ultimately proposing {\em sticky actions} as a standard training and evaluation protocol, which will be incorporated in a new version of the Arcade Learning Environment. 

\subsection{The Brute}

The Brute is an algorithm designed to exploit features of the original Arcade Learning Environment. Although developed independently by some of this article's authors, it shares many similarities with the trajectory tree method of \citet{kearns99approximate}. The Brute uses the agent's trajectory $h_t = a_1, o_1, a_2, o_2, \ldots, o_t$ as state representation, assigning individual values to each state. Because of the ALE's determinism, a single sample from each state-action pair is sufficient for a perfect estimate of the agent's return up to that point. The Brute maintains a partial history tree that contains all visited histories. Each node, associated with a history, maintains an action-conditional transition function and a reward function. The Brute estimates the value for any history-action pair using bottom-up dynamic programming. The agent follows the best trajectory found so far, with infrequent random actions used to search for better trajectories.

In order to be able to apply the Brute to stochastic environments, our implementation maintains the maximum likelihood estimate for both transition and reward functions. We provide a full description of the Brute in Appendix~\ref{sec:brute}.

\subsubsection{Empirical Evaluation}
\label{sec:eval_determinism}

We evaluated the performance of the Brute on the five training games proposed by \citeauthor{Bellemare_JAIR13}~\citeyear{Bellemare_JAIR13}. The average score obtained by the Brute, as well as of DQN and Sarsa($\lambda$) $+$ Blob-PROST, are presented in Table~\ref{tab:eval_deterministic_brute}. Agents interacted with the environment for 50 million frames and the numbers reported are the average scores agents obtained in the last 100 episodes played while learning. We discuss our experimental setup in Appendix~\ref{sec:experimental_setup}.

\begin{table}[t]
\centering
\caption{The Brute's performance compared to Sarsa($\lambda$) $+$ Blob-PROST and DQN in the deterministic Arcade Learning Environment. Standard deviation over trials is reported between parenthesis. See text for details.}
\scriptsize{
  \begin{tabular}{ l | r l | r l | r l }
    Game  &\multicolumn{2}{| c |}{The Brute} &\multicolumn{2}{| c |}{Sarsa($\lambda$) $+$ Blob-PROST} &\multicolumn{2}{| c }{DQN}\\
    \hline \hline
    \textsc{Asterix}        &6,909  &(1,018)   &\ \ \ \ \ 4,173   &(872)      &3,501       &(420)      \\ \hdashline[0.5pt/2pt]
    \textsc{Beam Rider}     &1,132  &(178)     &\ \ \ \ \ 2,098   &(508)      &4,687       &(704)      \\ \hdashline[0.5pt/2pt]
    \textsc{Freeway}        &1.1    &(0.4)     &\ \ \ \ \ 32.1    &(0.4)      &32.2        &(0.1)      \\ \hdashline[0.5pt/2pt]
    \textsc{Seaquest}       &621    &(192)     &\ \ \ \ \ 1,340   &(245)      &1,397       &(215)      \\ \hdashline[0.5pt/2pt]
    \textsc{Space Invaders} &1,432  &(226)     &\ \ \ \ \ 723     &(86)       &673         &(18)       \\
  \end{tabular}
}
\label{tab:eval_deterministic_brute}
\end{table}

The Brute is crude but we see that it leads to competitive performance in a number of games. In fact, \citet{Bellemare_IJCAI15}, using a different evaluation protocol, reports that the Brute outperformed the best learning method at the time on 45 out of 55 Atari 2600 games. However, as we will see, this performance critically depends on the environment's determinism. In the next section we discuss how we modified the ALE to introduce a form of stochasticity we call sticky actions; and we show that the Brute fails when small random perturbations are introduced.

\subsection{Sticky Actions}

This section introduces {\em sticky actions}, our approach to injecting stochasticity into the ALE. This approach also evaluates the robustness of learned policies. Its design is based on the following desiderata:
\begin{itemize}
      \item the stochasticity should be minimally non-Markovian with respect to the environment, \emph{i.e.}, the action to be executed by the emulator should be conditioned only on the action chosen by the agent and on the previous action executed by the emulator,
      \item the difficulty of existing tasks should not be changed, \emph{i.e.}, algorithms that do not rely on the environment's determinism should not have their performance hindered by the introduction of stochasticity, and
      \item it should be easy to implement in the ALE, not requiring changes inside the Stella emulator, but only on the framework itself.
\end{itemize}

In sticky actions there is a {\em stickiness} parameter $\varsigma$, the probability at every time step that the environment will execute the agent's previous action again, instead of the agent's new action. More specifically, at time step $t$ the agent decides to execute action $a$; however, the action $A_t$ that the environment in fact executes is:

$$A_t = 
      \left\{\begin{array}{rl} 
            a, &\mbox{with prob.}\quad 1 - \varsigma,\\
            a_{t-1}, &\mbox{with prob.}\quad \varsigma.
      \end{array}\right.
$$ 

\begin{figure*}
\center{
\includegraphics[width=4in]{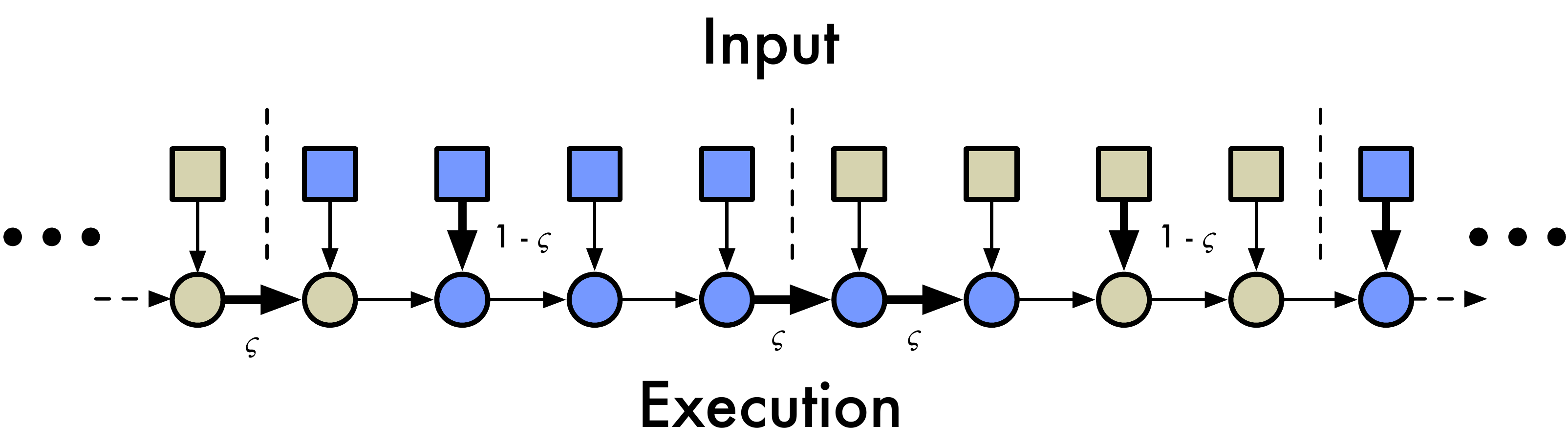}
\includegraphics[width=1.5in]{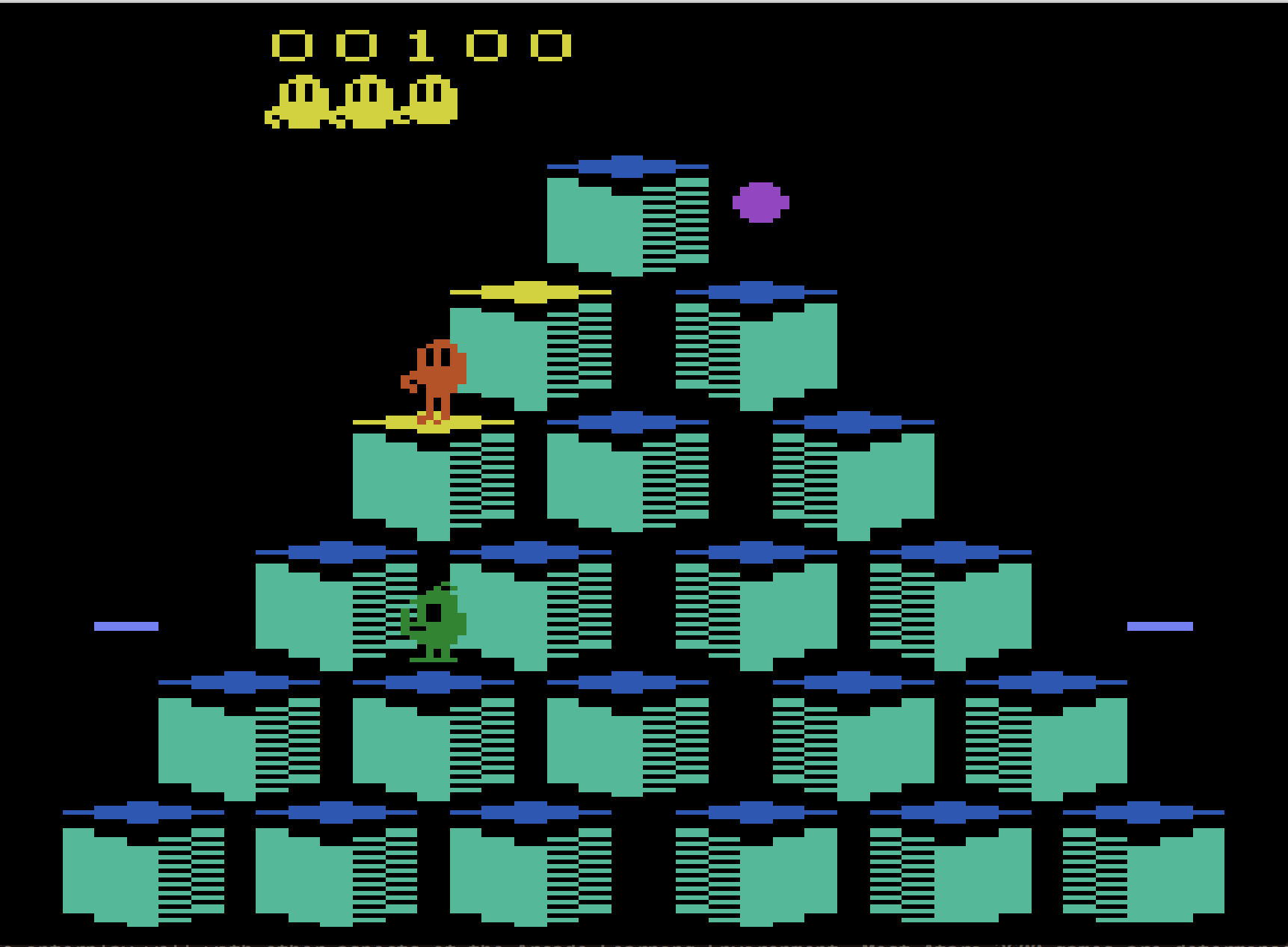}
}
\caption{\textbf{Left.} Interaction between the environment's input and the action it executes. Different colors represent different actions, boldface arrows indicate time steps at which past execution and input disagree. With probability $\varsigma$, the agent's input is ignored and the immediately preceding action is instead repeated. Vertical dotted lines indicate frame skipping boundaries; note that these are for illustration only, as our approach does not depend on frame skipping. \textbf{Right.} \gamename{Q*bert} is one game where different stochasticity models have significantly different effects.\label{fig:stochasticity_diagram}}
\end{figure*}

\noindent In other words, if $\varsigma = 0.25$, there is 25\% chance the environment will not execute the desired action right away. Figure \ref{fig:stochasticity_diagram} (left) illustrates this process.

Notice that if an agent decides to select the same action for several time steps, the time it will take to have this action executed in the environment follows a geometric distribution. The probability the previous action is executed $k$ times before the new action is executed is $(1 - \varsigma)^k \varsigma$. 

Sticky actions are different from random delays because, in the former, the agent can change its mind at any time by sending a new action to the emulator. To see why this matters, consider the game \gamename{Q*bert}, where a single wrong action may cause the agent to jump off the pyramid and lose a life (Figure \ref{fig:stochasticity_diagram}, right). Under sticky actions, the agent can switch to a no-op before landing on the edge, knowing that with high probability the action will not be continued up to the point it pushes the agent off the pyramid. With random delays, the previous action will be executed until the delay is passed, even if the agent switched to a no-op before landing on the edge. This increases the likelihood the agent will be forced to continue moving once it lands on the edge, making it more likely to fall off the pyramid.

Sticky actions also interplay well with other aspects of the Arcade Learning Environment. Most Atari 2600 games are deterministic and it would be very hard to change their dynamics. Our approach only impacts which actions are sent to be executed. Sticky actions  also interacts well with frame skipping (\emph{c.f.} Section~\ref{sec:background}). With sticky actions, at each intermediate time step between the skipped frames there is a probability $\varsigma$ of executing the previous action. Obviously, this applies until the current action is executed, when the previous action taken and the current action become the same. Figure~\ref{fig:stochasticity_diagram} depicts the process for a frame skip of 4. 

\subsubsection{Evaluating the Impact of Sticky Actions}

We now re-evaluate the performance of the Brute, DQN and Sarsa($\lambda$) $+$ Blob-PROST under the sticky actions protocol. The intuition is that the Brute, which exploits the assumption that the environment is deterministic, should perform worse when stochasticity is introduced. We repeated the experiments from Section~\ref{sec:eval_determinism}, but with $\varsigma = 0.25$. Table~\ref{tab:eval_brute} depicts the algorithms' performance in both the stochastic environment and in the deterministic environment.

\begin{table}[t]
\centering
\caption{The impact of stochasticity on different algorithms. Average over several trials is reported. Standard deviation over trials is within parentheses. The deterministic setting uses $\varsigma = 0.0$ while the stochastic setting uses $\varsigma = 0.25$. See text for details.}
\scriptsize{
  \begin{tabular}{ l | r l ;{0.5pt/2pt} r l | r l ;{0.5pt/2pt} r l | r l ;{0.5pt/2pt} r l }
    \multirow{2}{*}{Game}  &\multicolumn{4}{| c |}{The Brute} &\multicolumn{4}{| c |}{Sarsa($\lambda$) $+$ Blob-PROST} &\multicolumn{4}{| c }{DQN}\\
    &\multicolumn{2}{| c ;{0.5pt/2pt}}{Determin.} &\multicolumn{2}{;{0.5pt/2pt} c |}{Stochast.} &\multicolumn{2}{| c ;{0.5pt/2pt}}{Determin.} &\multicolumn{2}{;{0.5pt/2pt} c |}{Stochast.} &\multicolumn{2}{| c ;{0.5pt/2pt}}{Determin.} &\multicolumn{2}{;{0.5pt/2pt} c }{Stochast.}\\ \hline \hline
    \textsc{Asterix}        &6909 &(1018)     &308 &(31)      &4173   &(872)  &3411    &(414)      &3501 &(420)     &3123  &(96)   \\ \hdashline[0.5pt/2pt]
    \textsc{Beam Rider}     &1132 &(178)      &428 &(18)      &2098   &(508)  &1851    &(407)      &4687 &(704)     &4552  &(849)  \\ \hdashline[0.5pt/2pt]
    \textsc{Freeway}        &1.1   &(0.4)     &0.0 &(0.0)     &32.1   &(0.4)  &31.8    &(0.3)      &32.2 &(0.1)     &31.6  &(0.7)  \\ \hdashline[0.5pt/2pt]
    \textsc{Seaquest}       &621   &(192)     &81  &(7)       &1340   &(245)  &1204    &(190)      &1397 &(215)     &1431  &(162)  \\ \hdashline[0.5pt/2pt]
    \textsc{Space Invaders} &1432 &(226)      &148 &(11)      &723    &(86)   &583     &(31)       &673  &(18)      &687   &(37)   \\
  \end{tabular}
}
\label{tab:eval_brute}
\end{table}

We can see that the Brute is the only algorithm substantially impacted by the sticky actions. These results suggest that sticky actions enable us to empirically evaluate an agent's robustness to perturbation.

\vspace{3cm}

\subsection{Alternative Forms of Stochasticity}~\label{subsec:alternatives}

To conclude this section, we briefly discuss some alternatives to sticky actions, listing their pros ($+$) and cons ($-$). These alternatives fall in two broad categories: start-state methods and stochastic methods. In start-state methods, the first state of an episode is chosen randomly, but the deterministic dynamics remain unchanged. These approaches are less intrusive as the agent retains full control over its actions, but do not preclude exploiting the environment's determinism. This may be undesirable in games where the agent can exploit game bugs by executing a perfectly timed sequence of actions, as in, for example, the game \gamename{Q*bert}.\footnote{Personal communication from Ziyu Wang.} On the other hand, stochastic methods impact the agent's ability to control the environment uniformly throughout the episode, and thus its performance. We believe our proposed method minimizes this impact. \\

\textbf{Initial no-ops.} When evaluating the agent, begin the episode by taking from 0 to $k$ no-op actions, selected uniformly at random \citep{Mnih_Nature15}. By affecting the initial emulator state, this prevents the simplest form of open-loop control.
\begin{itemize}
      \item[$+$] No interference with agent action selection.
      \item[$-$] Impact varies across games. For example, initial no-ops have no effect in \gamename{Freeway}.
      \item[$-$] The environment remains deterministic beyond the choice of starting state.
      \item[$-$] Brute-like methods still perform well.
\end{itemize}

\textbf{Random human starts.} When evaluating the agent, randomly pick one of $k$ predetermined starting states. \shortciteauthor{Nair_Workshop15}~\citeyear{Nair_Workshop15}, for example, sampled starting states at random from a human's gameplay.
\begin{itemize}
      \item [$+$] Allows evaluating the agent in very different situations.
      \item[$-$] The environment remains deterministic beyond the choice of starting state.
      \item[$-$] Brute-like methods still perform well.
      \item [$-$] It may be difficult to provide starting states that are both meaningful and free of researcher bias. For example, scores as reported by \citeauthor{Nair_Workshop15}~\citeyear{Nair_Workshop15}  are not comparable across starting states: although in a full game of \gamename{Pong} an agent can score 21 points, from a much later starting state this score is unachievable.
\end{itemize}

\textbf{Uniformly random action noise.} With a small probability $\varsigma$, the agent's selected action is replaced with another action drawn uniformly from the set of legal actions. 
\begin{itemize}
      \item[$+$] Matches the most commonly used form of exploration, $\epsilon$-greedy. 
      \item[$-$] May significantly interfere with agent's policy, \emph{e.g.}, when navigating a narrow cliff such as in the game \gamename{Q*bert}.
\end{itemize}

\textbf{Random frame skips.} This approach, implemented in OpenAI's Gym \shortcite{brockman16openai}, is closest to our method. Each action randomly lasts between $k_1$ and $k_2$ frames.
\begin{itemize}
      \item[$+$] Does not interfere with action selection, only the timing of action execution.
      \item[$-$] This restricts agents to using frame skip. In particular, the agent cannot react to events occurring during an action's period.
      \item[$-$] Discounting must also be treated more carefully, as this makes the effective discount factor random.
      \item[$-$] The agent has perfect reaction time since its actions always have an immediate effect.
\end{itemize}

\textbf{Asynchronous environment.} More complex environments might involve unpredictable communication delays between the agent and the environment. This is the case in Minecraft (Project Malmo; \shortciteauthor{johnson16malmo}, 2016), Starcraft \shortcite{ontanon13survey}, and robotic RL platforms \shortcite{sutton11horde}. 
\begin{itemize}
      \item[$+$] This setting naturally discourages agents relying on determinism.
      \item[$-$] Lacks reproducibility across platforms and hardware.
      \item[$-$] With sufficiently fast communications, reverts to a deterministic environment.
\end{itemize}

\textbf{Overall comparison.} Our proposed solution, sticky actions, leverages some of the main benefits of other approaches without most of their drawbacks. It is free from researcher bias, it does not interfere with agent action selection, and it discourages agents from relying on memorization. The new environment is stochastic for the whole episode, generated results are reproducible, and our approach interacts naturally with frame skipping and discounting.

\section{Benchmark Results in the Arcade Learning Environment}
\label{sec:benchmark}

In this section we introduce new benchmark results for DQN and Sarsa($\lambda$) $+$ Blob-PROST in 60 different Atari 2600 games using sticky actions. It is our hope that future work will adopt the experimental methodology described in this paper, and thus be able to directly compare results with this benchmark.

\subsection{Experimental Method}

We evaluated DQN and Sarsa($\lambda$) $+$ Blob-PROST in 60 different Atari 2600 games. We report results using the sticky actions option in the new version of the ALE ($\varsigma = 0.25$), evaluating the final performance while learning, at 10, 50, 100 and 200 million frames. We computed score averages of each trial using the 100 final episodes until the specified threshold, including the episode in which the total is exceeded. We report the average over 5 trials for DQN and the average over 24 trials for Sarsa($\lambda$) $+$ Blob-PROST. To ease reproducibility, we listed all the relevant parameters used by Sarsa($\lambda$) $+$ Blob-PROST and DQN in Appendix~\ref{sec:experimental_setup}. We encourage researchers to present their results on the ALE in the same reproducible fashion.

\subsection{Benchmark Results}
\label{Results}

We present excerpts of the obtained results for Sarsa($\lambda$) $+$ Blob-PROST and DQN in Tables~\ref{tab:summary_blob_prost} and \ref{tab:summary_dqn}. These tables report the obtained scores in the games we used for training. These games were originally proposed by \citet{Bellemare_JAIR13}. The complete results are available in Appendix~\ref{sec:full_benchmark_results}.

\begin{table}[t]
\centering
\caption{Results on the ALE's original training set using Sarsa($\lambda$) $+$ Blob-PROST. Averages over 24 trials are reported and standard deviation over trials is presented between parenthesis.}
\scriptsize{
  \begin{tabular}{ l | r l ;{0.5pt/2pt} r l ;{0.5pt/2pt} r l ;{0.5pt/2pt} r l }
    Game &\multicolumn{2}{| c ;{0.5pt/2pt}}{10M frames} &\multicolumn{2}{;{0.5pt/2pt} c ;{0.5pt/2pt}}{50M frames} &\multicolumn{2}{;{0.5pt/2pt} c ;{0.5pt/2pt}}{100M frames} &\multicolumn{2}{;{0.5pt/2pt} c }{200M frames}\\
    \hline \hline
    \textsc{Asterix}         &2,088.3 &(302.5) &3,411.0  &(413.5) &3,768.1  &(312.5)   &4,395.2  &(460.7)   \\ \hdashline[0.5pt/2pt]
    \textsc{Beam Rider}      &1,149.1 &(235.2) &1,851.2  &(406.7) &2,116.4  &(516.0)   &2,231.9  &(470.5)   \\ \hdashline[0.5pt/2pt]
    \textsc{Freeway}         &28.7    &(5.1)   &31.8     &(0.3)   &31.9     &(0.2)     &31.8     &(0.2)     \\ \hdashline[0.5pt/2pt]
    \textsc{Seaquest}        &747.9   &(222.2) &1,204.2  &(189.8) &1,327.1  &(337.9)   &1,403.1  &(301.7)   \\ \hdashline[0.5pt/2pt]
    \textsc{Space Invaders}  &458.2   &(23.8)  &582.9    &(30.7)  &661.6    &(51.4)    &759.7    &(43.9)
  \end{tabular}
}
\label{tab:summary_blob_prost}
\end{table}

Because we report the algorithms' performance at different points in time, these results give us insights about learning progress made by each algorithm. Such analysis allows us to verify, across 60 games, how often an agent's performance plummets; as well as how often agents reach their best performance before 200 million frames.

In most games, Sarsa($\lambda$) $+$ Blob-PROST's performance steadily increases for the whole learning period. In only 10\% of the games the scores obtained with 200 million frames are lower than the scores obtained with 100 million frames. This difference is statistically significant in only 3 games:\footnote{\label{footnote_sarsa}Welch's t-test ($p < 0.05$; $n = 24$).} \textsc{Carnival}, \textsc{Centipede}, and \textsc{Wizard of Wor}. However, in most games we observe diminishing improvements in an agent's performance. In only 22 out of 60 games we observe statistically significant improvements from 100 million frames to 200 million frames.\textsuperscript{\ref{footnote_sarsa}} In several games such as \textsc{Montezuma's Revenge} this stagnation is due to exploration issues; the agent is not capable of finding additional rewards in the environment.

DQN has much higher variability in the learning process and it does not seem to benefit much from additional data. DQN obtained its highest scores using 200 million frames in only 35 out of 60 games. Agents' performance at 200 million frames was statistically better than agents' performance at 100 million frames in only 18 out of 60 games.\footnote{\label{footnote_dqn}Welch's t-test ($p < 0.05$; $n = 5$).} In contrast, Sarsa($\lambda$) $+$ Blob-PROST achieves its highest scores with 200 million samples in 50 out of 60 games. We did not observe statistically significant performance decreases for DQN when comparing agents' performance at 100 and 200 million samples.\textsuperscript{\ref{footnote_dqn}} It is important to add a caveat that the lack of statistically significant results may be due to our sample size ($n = 5$). The t-test's power may still be too low to detect significant differences in DQN's performance. It is worth pointing out that when DQN was originally introduced, its results consisted of only one independent trial. Despite its high computational cost we evaluated it on 5 trials in an attempt to evaluate such an important algorithm more thoroughly, addressing the methodological concerns we discussed above and offering a more reproducible and statistically comparable DQN benchmark.

We also compared the performance of both algorithms in each game to understand specific trends such as performance plumetting and absence of learning. Performance drops seem to be algorithm dependent, not game dependent. \textsc{Centipede} is the only game in which plummeting performance was observed for both both DQN and Sarsa($\lambda$) $+$ Blob-PROST. The decrease in performance we observe in other games occurs only for one algorithm. On the other hand, we were able to identify some games that seem to be harder than others for both algorithms. Both algorithms fail to make much progress on games such as \textsc{Asteroids}, \textsc{Pitfall}, and \textsc{Tennis}. These games generally pose hard exploration tasks to the agent; or have complex dynamics, demanding better representations capable of accurately encoding value function approximations.

\begin{table}[t]
\centering
\caption{Results on the ALE's original training set using DQN. Averages over 5 trials are reported and standard deviation over trials is presented between parenthesis.}
\scriptsize{
  \begin{tabular}{ l | r l ;{0.5pt/2pt} r l ;{0.5pt/2pt} r l ;{0.5pt/2pt} r l }
    Game &\multicolumn{2}{| c ;{0.5pt/2pt}}{10M frames} &\multicolumn{2}{;{0.5pt/2pt} c ;{0.5pt/2pt}}{50M frames} &\multicolumn{2}{;{0.5pt/2pt} c ;{0.5pt/2pt}}{100M frames} &\multicolumn{2}{;{0.5pt/2pt} c }{200M frames}\\
    \hline \hline
    \textsc{Asterix}         &1,732.6 &(314.6)   &3,122.6   &(96.4)    &3,423.4 &(213.6) &2,866.8 &(1,354.6) \\ \hdashline[0.5pt/2pt]
    \textsc{Beam Rider}      &693.9   &(111.0)   &4,551.5   &(849.1)   &4,977.2 &(292.2) &5,700.5 &(362.5)   \\ \hdashline[0.5pt/2pt]
    \textsc{Freeway}         &13.8    &(8.1)     &31.7      &(0.7)     &32.4    &(0.3)   &33.0    &(0.3)     \\ \hdashline[0.5pt/2pt]
    \textsc{Seaquest}        &311.5   &(36.9)    &1,430.8   &(162.3)   &1,573.4 &(561.4) &1,485.7 &(740.8)   \\ \hdashline[0.5pt/2pt]
    \textsc{Space Invaders}  &211.6   &(14.8)    &686.6     &(37.0)    &787.2   &(173.3) &823.6   &(335.0)   \\ \hdashline[0.5pt/2pt]
  \end{tabular}
}
\label{tab:summary_dqn}
\end{table}

We can also compare our results to previously published results to verify the impact our proposed evaluation protocol has in agents' performance. This new setting does not seem to benefit a specific algorithm. Sarsa($\lambda$) $+$ Blob-PROST and DQN still present comparable performance, with each algorithm being better in an equal number of games, as suggested by \citet{Liang_AAMAS16}. As we already discussed in Section~\ref{sec:determinism_stochasticity}, using sticky actions seems to only substantially hinder the performance of the Brute agent, not having much impact in the performance of DQN and Sarsa($\lambda$) $+$ Blob-PROST. We observed decreased performance for DQN and Sarsa($\lambda$) $+$ Blob-PROST only in three games: \textsc{Breakout}, \textsc{Gopher}, and \textsc{Pong}.

\section{Open Problems and the Current State-of-the-Art in the ALE}
\label{sec:open_prob}

To provide a complete big picture of how the ALE is being used by the research community. It is also important to discuss the variety of research problems for which the community has used the ALE as a testbed. In the past few years we have seen several successes showcased in the ALE, with new results introduced at a rapid pace.

We list five important research directions the community has worked on using the ALE, and we use current results in the literature to argue that while there has been substantial progress these problems still remain open. These research directions are:
\begin{itemize}
\item representation learning,
\item exploration, 
\item transfer learning, 
\item model learning, and
\item off-policy learning.
\end{itemize}

\subsection{Representation Learning}

The ALE was originally introduced to pose the problem of general competency: expecting a single algorithm to be capable of playing dozens of Atari 2600 games. Therefore, agents must either use generic encodings capable of representing all games (\emph{e.g.}, \citeauthor{Liang_AAMAS16}, 2016), or be able to automatically learn representations. The latter is obviously more desirable for the potential of discovering better representations while alleviating the burden of having handcrafted features. 

\begin{figure}
  \centering
  \begin{subfigure}[b]{0.4\textwidth} 
    \includegraphics[width=\textwidth]{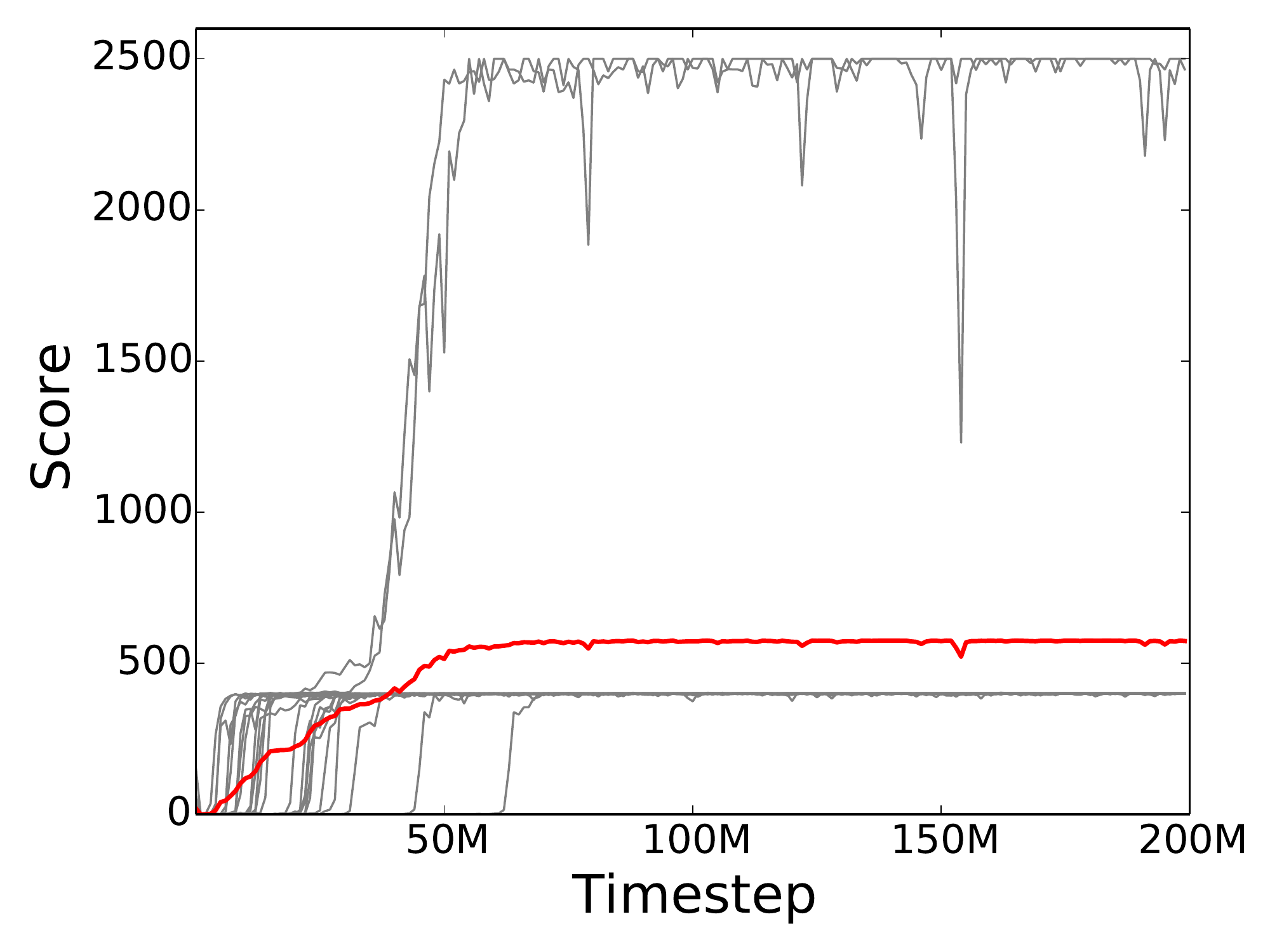} 
    \centering
    \caption{Sarsa($\lambda$) $+$ Blob-PROST} 
    \label{fig:montezumaSarsa} 
  \end{subfigure}
  \begin{subfigure}[b]{0.4\textwidth} 
    \includegraphics[width=\textwidth]{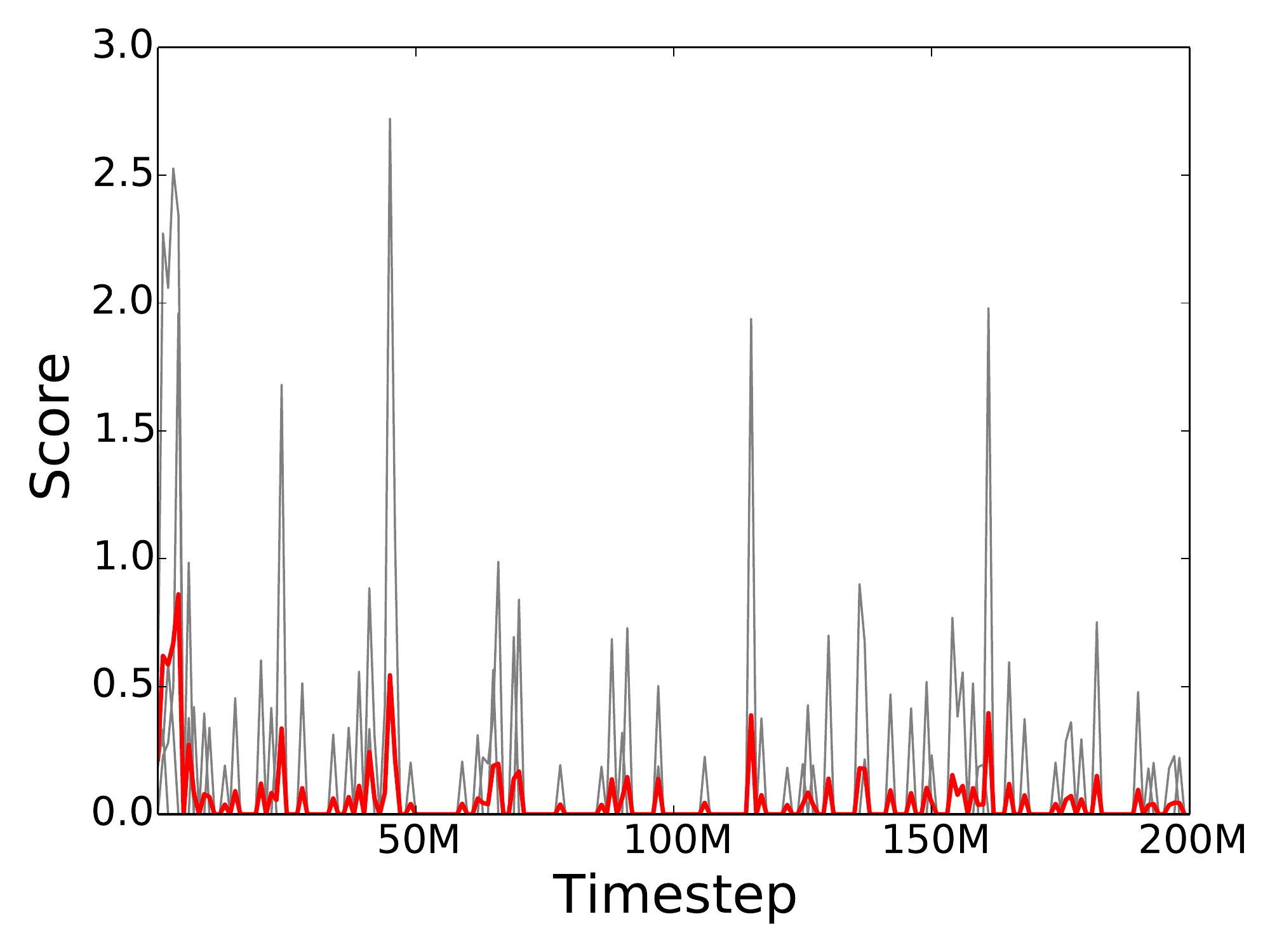}
    \centering
    \caption{DQN}
    \label{fig:montezumaDQN}
  \end{subfigure}
  \caption{Comparison between learning curves of DQN and Sarsa($\lambda$) $+$ Blob-PROST in \textsc{Montezuma's Revenge}. Notice the y-axes are not on the same scale. Each point corresponds to the average performance over the last one million episodes. Grey curves depict individual trials. The red curve depicts the average over all trials.}
  \label{fig:comparisonMontezuma}
\end{figure}

Deep Q-Networks (DQN) of \citeauthor{Mnih_Nature15}~\citeyear{Mnih_Nature15} demonstrate it is possible to learn representations jointly with control policies. However, reinforcement learning methods based on neural networks still have a high sample complexity, requiring at least dozens of millions of samples before achieving good performance, in part due to the need for learning this representation. In the results we report, DQN's performance (Table~\ref{tab:benchmark_dqn}) is better than Sarsa($\lambda$) $+$ Blob-PROST's (Table~\ref{tab:benchmark_blob_prost}) in less than 20\% of the games when evaluated at 10 million frames, and achieves comparable performance at 100 million frames. The high sample complexity also seems to hinder the agents' performance in specific environments, such as when non-zero rewards are very sparse. Figure~\ref{fig:comparisonMontezuma} illustrates this point by showing how DQN sees non-zero rewards occasionally while playing \textsc{Montezuma's Revenge} (Figure~\ref{fig:montezuma_revenge}), but it does not learn to obtain non-zero rewards consistently. Recently, researchers have tried to address this issue by weighting samples differently, prioritizing those that seem to provide more information to the agent~\cite{Schaul_ICLR16}. Another approach is to use auxiliary tasks that allow agents to start learning a representation before the first extrinsic reward is observed~\shortcite{jaderberg17reinforcement}; the distributions output by the C51 algorithm of \citet{bellemare17distributional} may be viewed as a particularly meaningful set of auxiliary tasks. Finally, intrinsically generated rewards \citep{Bellemare_NIPS16} may also provide a useful learning signal which the agent can use to build a representation. 

Despite this high sample complexity, DQN and DQN-like approaches remain the best performing methods overall when compared to simple, hand-coded representations~\cite{Liang_AAMAS16}. However, these improvements are not as dramatic as they are in other applications (\emph{e.g.}, computer vision; \shortciteauthor{Krizhevsky12}, 2012). Furthermore, this superior performance often comes at the cost of additional tuning, as recently reported by \citet{islam17reproducibility} in the context of continuous control. This suggests that there is still room for significant progress on effectively learning good representations in the ALE.

Different approaches that learn an internal representation in a sample efficient way have also been proposed \shortcite{Veness_AAAI15}, although they have not yet been fully explored in this setting. Other directions the research community has been looking at are the development of better visualization methods~\cite{Zahavy16}, the proposal of algorithms that alleviate the need for specialized hardware~\cite{Mnih_ICML16}, and genetic algorithms \cite{kelly17emergent}. 

\subsection{Planning and Model-Learning}

Despite multiple successes of search algorithms in artificial intelligence (\emph{e.g.}, \shortciteauthor{Campbell_AI02}, 2002; \shortciteauthor{Schaeffer_Science07}, 2007; \shortciteauthor{Silver_Nature16}, 2016), planning in the Arcade Learning Environment remains rare compared to methods that learn policies or value functions ~(but see \shortciteauthor{Bellemare_JAIR13}, 2013b; \shortciteauthor{Guo_NIPS14}, 2014; \shortciteauthor{Lipovetzky_IJCAI15}, 2015; \shortciteauthor{Shleyfman_IJCAI16}, 2016; \shortciteauthor{Jinnai_AAAI17}, 2017, for published planning results in the ALE). Developing heuristics that are general enough to be successfully applied to dozens of different games is a challenging problem. The problem's branching factor and the fact that goals are sometimes thousands of steps ahead of the agent's initial state are also major difficulties.

\begin{figure}
  \centering
  \includegraphics[width=\textwidth]{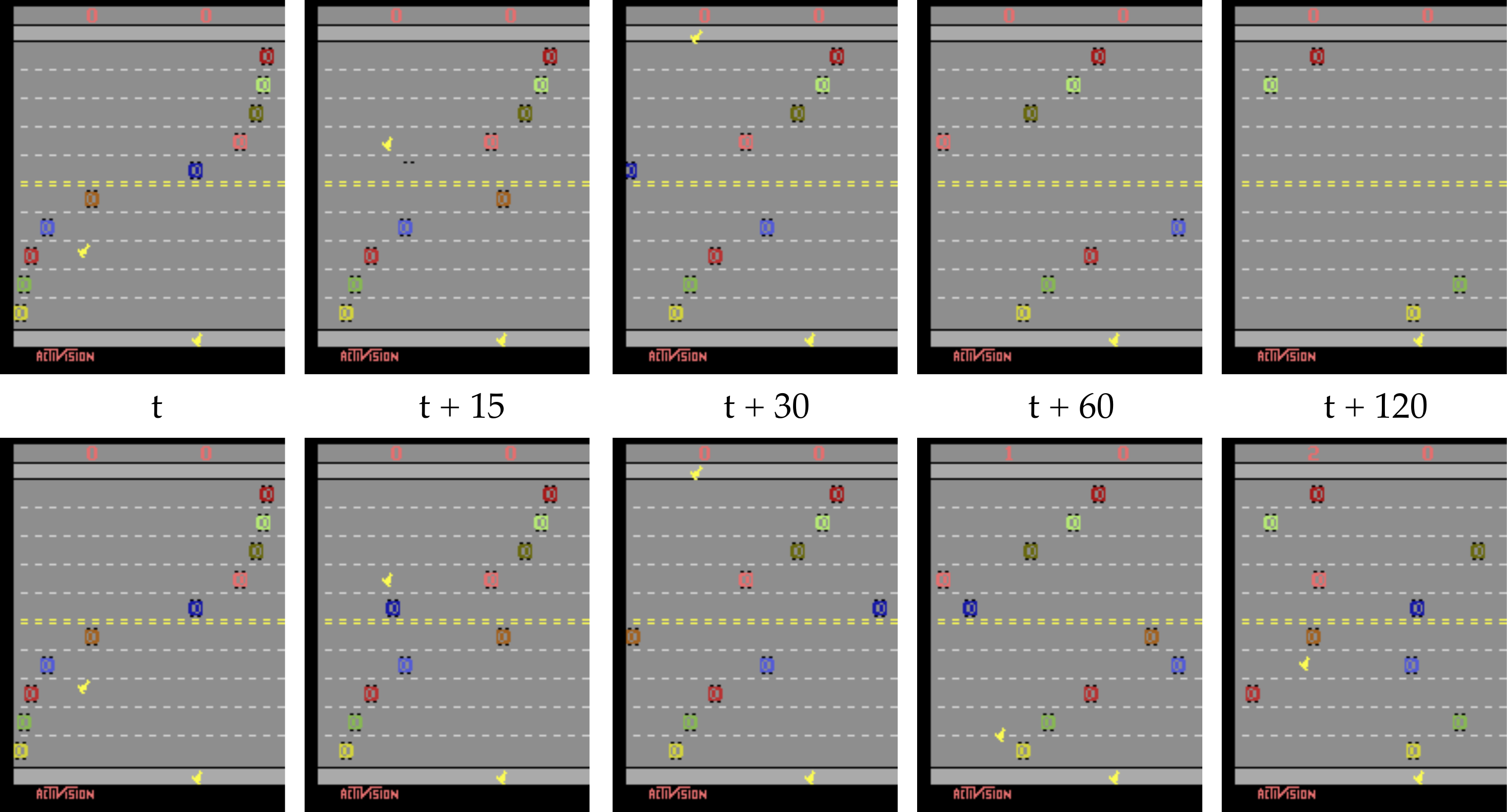}
  \caption{Top row: Rollout obtained with a learned model of the game \textsc{Freeway}. Bottom row: Ground truth. Small errors can be noticed ($t+15$) but major errors are observed only when the chicken crosses the street ($t + 30$), as depicted in frame $t + 60$. The score is not updated and the chicken does not respawn at the bottom of the screen. Later, cars start to disappear, as shown in the frame $t + 120$. This model was learned using quad-tree factorization~\cite{Bellemare_ICML13}.}
  \label{fig:ComparisonModelLearning}
\end{figure}

Almost all successes of planning in the ALE use the generative model provided by the Stella emulator, and so have an exact model of the environment. Learning generative models is a very challenging task~\shortcite{Bellemare_ICML13,Bellemare_ICML14,Oh_NIPS15,chiappa17recurrent} and so far, there has been no clear demonstration of successful planning with a learned model in the ALE. Learned models tend to be accurate for a small number of time steps until errors start to compound~\cite{Talvitie_UAI14}. As an example, Figure~\ref{fig:ComparisonModelLearning} depicts rollouts obtained with one of the first generative models trained on the ALE \citep{Bellemare_ICML13}. In this figure we can see how the accuracy of rollouts start to drop after a few dozen time steps. Probably the most successful example of model learning in the ALE is due to \shortciteauthor{Oh_NIPS15}~\citeyear{Oh_NIPS15} who learned multistep models that, up to one hundred time steps, appear accurate. These models are able to assist with exploration, an indication of the models' accuracy. However, because of compounding errors, the algorithm still needs to frequently restore its model to the real state of the game. More recently, \citet{chiappa17recurrent} showed significant improvements over this original model, including the ability to plan with the internal state. In both cases, however, the models are much slower than the emulator itself; designing a fast, accurate model remains an open problem. 

A related open problem is how to plan with an imperfect model. Although an error-free model might be unattainable, there is plenty of evidence that even coarse value functions are sufficient for the model-free case \citep{Veness_AAAI15}, raising the question of how to compensate for a model's flaws. Training set augmentation \citep{Talvitie_UAI14,Talvitie_AAAI17,venkatraman15improving} has shown that it is possible to improve an otherwise limited model. Similarly, \citet{farahmand16valueaware} showed that better planning performance could be obtained by using a value-aware loss function when training the model. We believe this to be a rich research direction.

\subsection{Exploration}

\begin{figure}[t]
  \centering
  \begin{subfigure}[b]{0.23\textwidth} 
    \includegraphics[width=\textwidth]{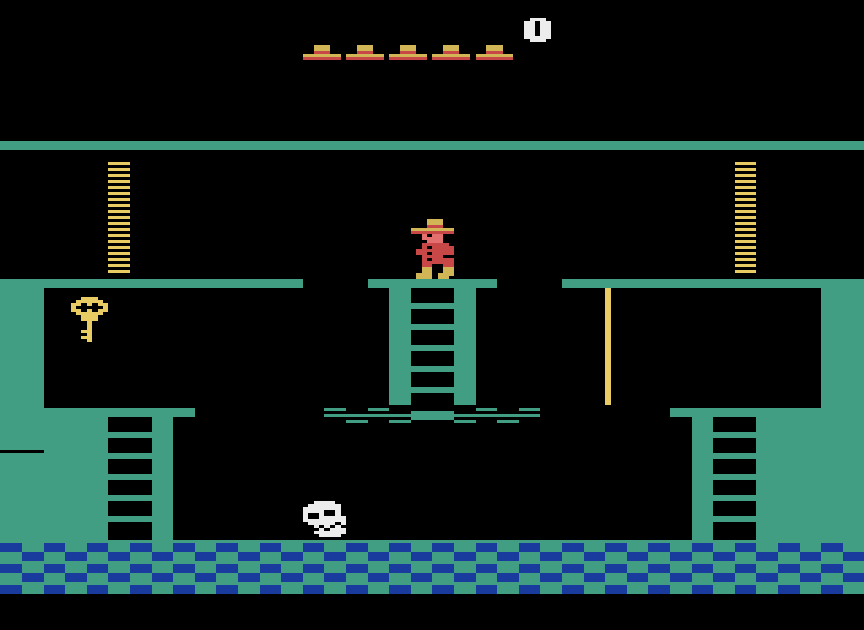} 
    \caption{\textsc{Mont. Revenge}} 
    \label{fig:montezuma_revenge} 
  \end{subfigure} ~ 
  \begin{subfigure}[b]{0.23\textwidth} 
    \includegraphics[width=\textwidth]{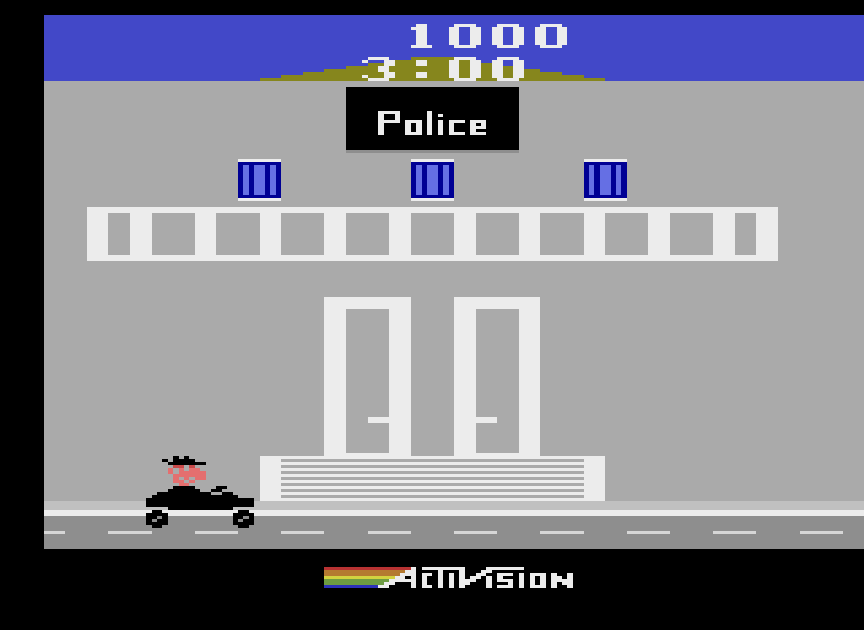}
    \caption{\textsc{Private Eye}}
    \label{fig:private_eye}
  \end{subfigure} ~
  \begin{subfigure}[b]{0.23\textwidth}
    \includegraphics[width=\textwidth]{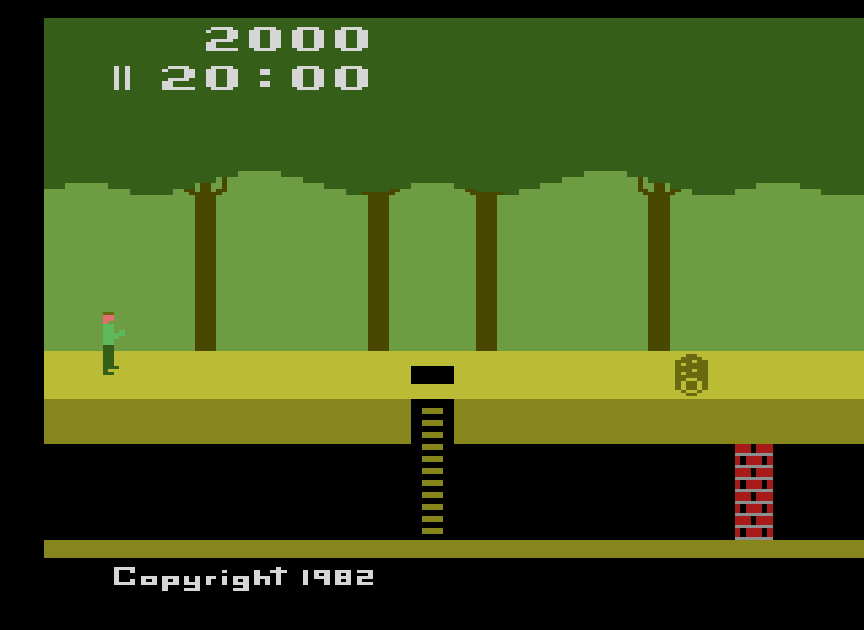}
    \caption{\textsc{Pitfall!}}
    \label{fig:pitfall}
  \end{subfigure}
  \begin{subfigure}[b]{0.23\textwidth}
    \includegraphics[width=\textwidth]{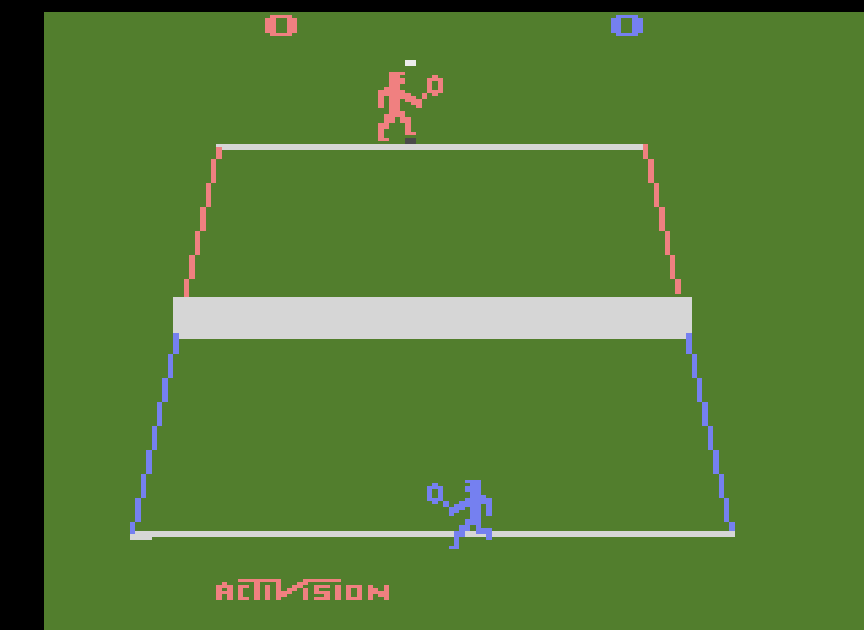}
    \caption{\textsc{Tennis}}
    \label{fig:tennis}
  \end{subfigure}  
  \caption{Challenging games in the ALE due to poor exploration.}
  \label{fig:hard_exploration}
\end{figure}

Most approaches for exploration focus on the tabular case and generally learn models of the environment (\emph{e.g.}, \citeauthor{Kearns_ML02}, 2002; \citeauthor{Brafman_JMLR02}, 2002; \citeauthor{Strehl_JCSS08}, 2008). The community is just beginning to investigate exploration strategies in model-free settings when function approximation is required~(\emph{e.g.,} \shortciteauthor{Bellemare_NIPS16}, 2016b; \shortciteauthor{Osband_NIPS16}, 2016; \shortciteauthor{Ostrovski17}, 2017; \shortciteauthor{Machado_ICML17}, 2017; \shortciteauthor{Martin_IJCAI17}, 2017; \shortciteauthor{Vezhnevets_ICML17}, 2017). This is the setting in which the ALE lies. Visiting every state does not seem to be a feasible strategy given the large number of possible states in a game (potentially $2^{1024}$ different states since the Atari 2600 has 1024 bits of RAM memory). In several games such as \textsc{Montezuma's Revenge} and \textsc{Private Eye} (see Figure~\ref{fig:hard_exploration}) even obtaining any feedback is difficult because thousands of actions may be required before a first positive reward is seen. Given the usual sample constraints (200 million frames), random exploration is highly unlikely to guide the agent towards positive rewards. In fact, some games such as \textsc{Pitfall!} and \textsc{Tennis}~(see Figure~\ref{fig:hard_exploration}) pose an even harder challenge: random exploration is more likely to yield negative rewards than positive ones.
In consequence, many simpler agents learn that staying put is the myopically best policy, although recent state-of-the-art agents (\emph{e.g.}, \citeauthor{bellemare17distributional}, 2017; \citeauthor{jaderberg17reinforcement}, 2017) can sometimes overcome this negative reward gradient. 

Some researchers recently started trying to address the exploration problem in the ALE. \shortciteauthor{Machado_LGCVG15}~\citeyear{Machado_LGCVG15} extended optimistic initialization to function approximation. \shortciteauthor{Oh_NIPS15}~\citeyear{Oh_NIPS15} and \shortciteauthor{Stadie_ARXIV15}~\citeyear{Stadie_ARXIV15} learned models to predict which actions lead the agent to frames observed least often, or with more uncertainty. \shortciteauthor{Bellemare_NIPS16}~\citeyear{Bellemare_NIPS16}, \citet{Ostrovski17} and \citet{Martin_IJCAI17} extended state visitation counters to the case of function approximation. \citet{Osband_NIPS16} uses randomized value functions to better explore the environment. \citet{Machado_ICML17} and \citet{Vezhnevets_ICML17} proposed the use of options to generate decisive agents, avoiding the dithering commonly observed in random walks. However, despite successes in individual games, such as \citeauthor{Bellemare_NIPS16}'s success in \textsc{Montezuma's Revenge}, none of these approaches has been able to improve, in a meaningful way, agents' performance in games such as \textsc{Pitfall!}, where the only successes to date involve some form of apprenticeship (\emph{e.g.}, \shortciteauthor{hester17learning},~2017).

There is still much to be done to narrow the gap between solutions applicable to the tabular case and solutions applicable to the ALE. An aspect that still seems to be missing are agents capable of committing to a decision for extended periods of time, exploring in a different level of abstraction, something that humans frequently do. Maybe agents should not be exploring in terms of joystick movements, but in terms of object configurations and game levels. Finally, for intrinsically difficult games, agents may need some form of intrinsic motivation \citep{oudeyer07intrinsic,barto13intrinsic} to keep playing despite the apparent impossibility of scoring in the game.

\subsection{Transfer Learning}

\begin{figure}
  \centering
  \begin{subfigure}[b]{0.3\textwidth} 
    \includegraphics[width=\textwidth]{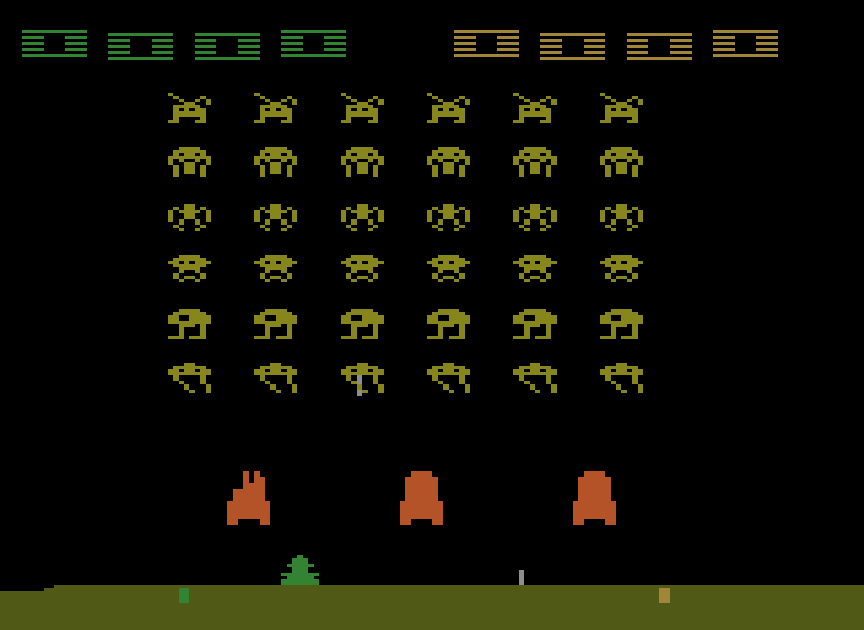} 
    \caption{\textsc{Space Invaders}} 
    \label{fig:space_invad
      ers} 
  \end{subfigure} ~ 
  \begin{subfigure}[b]{0.3\textwidth} 
    \includegraphics[width=\textwidth]{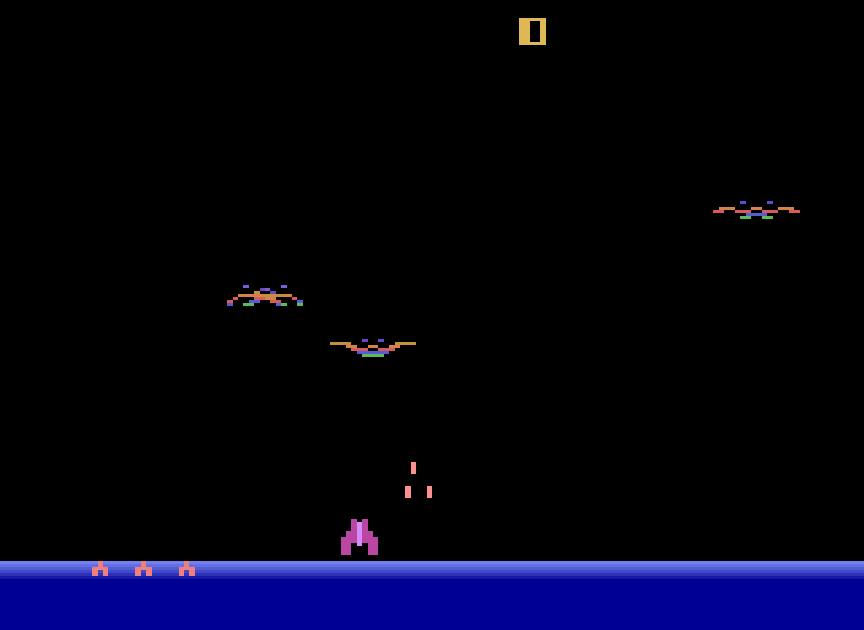}
    \caption{\textsc{Demon Attack}}
    \label{fig:demon_attack}
  \end{subfigure}
  \caption{Very similar games in the ALE.}
  \label{fig:similar_games}
\end{figure}

Most work in the ALE involves training agents separately in each game, but many Atari 2600 games have similar dynamics. We can expect knowledge transfer to reduce the required number of samples needed to learn to play similar games. As an example, \textsc{Space Invaders} and \textsc{Demon Attack} (Figure~\ref{fig:similar_games}) are two similar games in which the agent is represented by a spaceship at the bottom of the screen and it is expected to shoot incoming enemies. A more ambitious research question is how to leverage general video game experience, sharing knowledge across games that are not directly analogous. In this case, more abstract concepts could be learned, such as ``sometimes new screens are seen when the avatar goes to the edge of the current screen''.

There are attempts to apply transfer learning in the ALE~\shortcite{Rusu_ICLR16,Parisotto_ICLR16}. Such attempts are restricted to a dozen games that tend to be similar and generally require an ``expert'' network first, instead of learning how to play all games concurrently. \shortciteauthor{Taylor_JMLR_09}~\citeyear{Taylor_JMLR_09} have shown one can face \emph{negative transfer} depending on the similarity between the tasks being used. It is not clear how this should be addressed in the ALE. Ideally one would like to have an algorithm automatically deciding which games are helpful and which ones are not. Finally, current approaches are only based on the use of neural networks to perform transfer, conflating representation and policy transfer. It may be interesting to investigate how to transfer each one of these entities independently. To help explore these issues, the most recent version of the ALE supports game modes and difficulty settings.

\subsubsection{Modes and Difficulties in the Arcade Learning Environment}\label{sec:modes_diff_ale}

\begin{figure}
  \centering
  \begin{subfigure}[b]{0.3\textwidth} 
    \includegraphics[width=\textwidth]{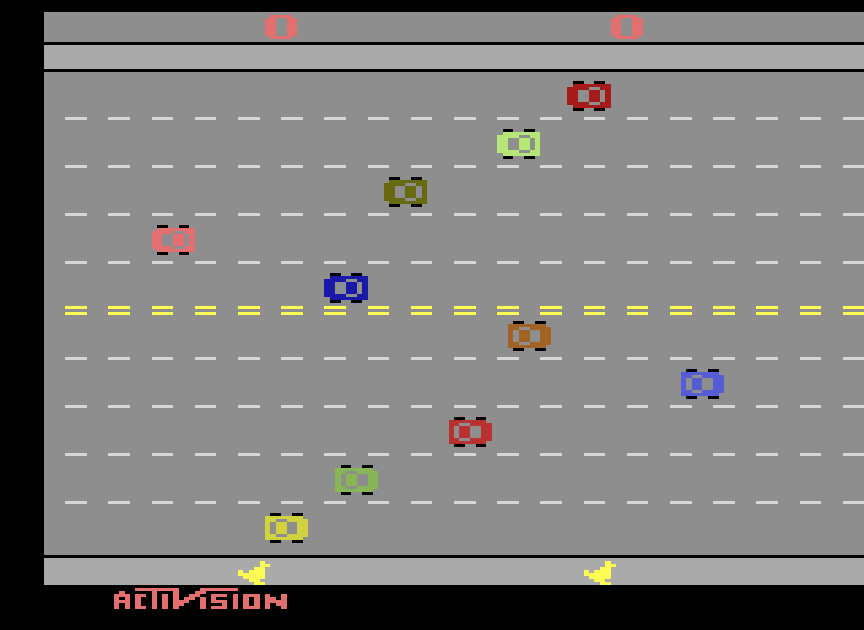} 
    \caption{Mode 1} 
    \label{fig:freeway_mode1} 
  \end{subfigure} ~ 
  \begin{subfigure}[b]{0.3\textwidth} 
    \includegraphics[width=\textwidth]{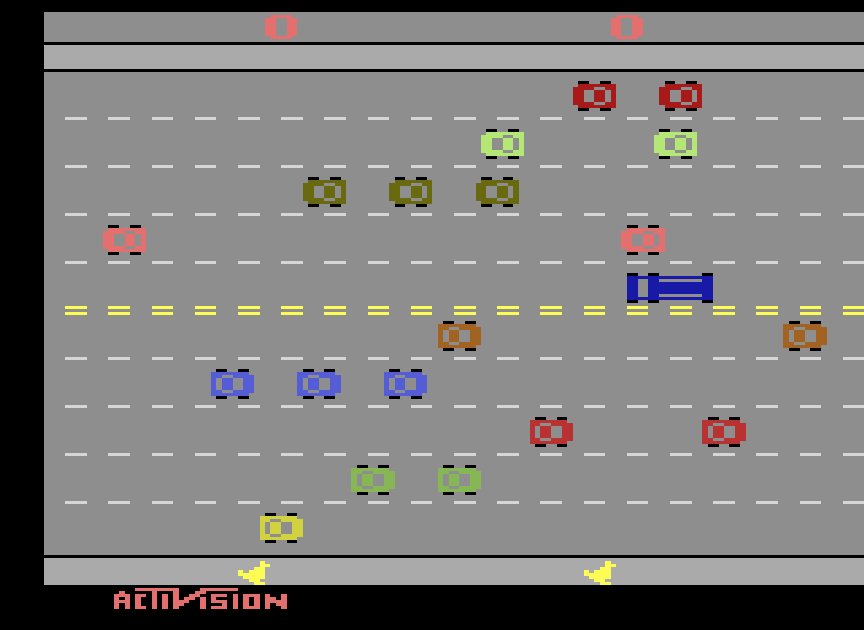}
    \caption{Mode 2}
    \label{fig:freeway_mode2}
  \end{subfigure} ~
  \begin{subfigure}[b]{0.3\textwidth}
    \includegraphics[width=\textwidth]{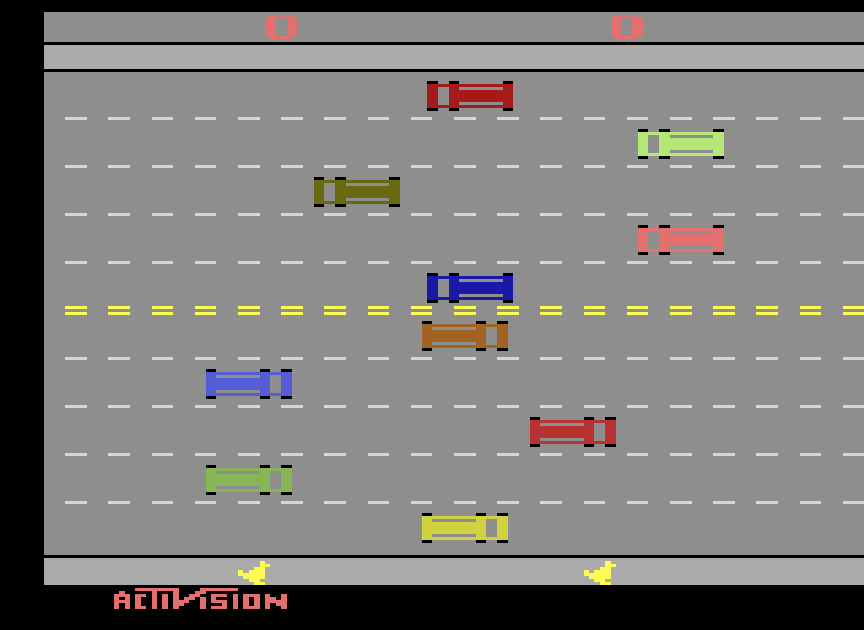}
    \caption{Mode 4}
    \label{fig:freeway_mode4}
  \end{subfigure}
  \caption{Different modes of the game \textsc{Freeway}.}
  \label{fig:freeway_modes}
\end{figure}

Originally, many Atari 2600 games had a default game mode and difficulty level that could be changed by select switches on the console. These mode/difficulty switches had different consequences such as changing the game dynamics or introducing new actions (see Figure~\ref{fig:freeway_modes}). Until recently, the ALE allowed agents to play games only in their default mode and difficulty. The newest version of the ALE allows one to select among all different game modes and difficulties that are single player games. We call each mode-difficulty pair a \emph{flavor}.

This new feature opens up research avenues by introducing dozens of new environments that are very similar. Because the underlying state representations across different flavors are probably highly related, we believe negative transfer is less likely, giving an easier setup for transfer. The list of such games the ALE will initially support, and their number of flavors, is available in Appendix~\ref{sec:flavors}. 

\subsection{Off-Policy Learning}

Off-policy learning algorithms seem to be brittle when applied to the ALE. \citeauthor{Defazio_ARXIV13}~\citeyear{Defazio_ARXIV13} have reported divergence when using algorithms such as GQ($\lambda$), without the projection step, and Q-learning, for instance. 

Besides the proposal of new algorithms that are theoretically better behaved~(\emph{e.g.}, \shortciteauthor{Maei_ICML10}, 2010), attempts to reduce divergence in off-policy learning currently consist of heuristics that try to decorrelate observations, such as the use of an experience replay buffer and the use of a target network in DQN~\cite{Mnih_Nature15}. Recent papers introduce changes in the update rules of Q-Learning to reduce overestimation of value functions~\shortcite{Hasselt_AAAI16}, new operators that increase the action-gap of value function estimates~\shortcite{Bellemare_AAAI16}, and more robust off-policy multi-step algorithms \shortcite{harutyunyan16qlambda,Munos_NIPS16}. However, besides a better theoretical understanding about convergence, stable (and practical) off-policy learning algorithms with function approximation are still an incomplete piece in the literature. So far, the best empirical results reported in the ALE were obtained with algorithms whose performance is not completely explained by current theoretical results. A thorough empirical evaluation of recent off-policy algorithms, such as GTD, remains lacking.

Addressing the aforementioned issues, either through a convincing demonstration of the efficacy of the current theoretically sound algorithms for off-policy learning, or through some of the improvements described above may free us from the increased complexity of using experience replay and/or target networks. Also, this would allow us to better reuse samples from policies that are very different from the one being learned.

\section{Conclusion}
\label{sec:conclusion}

In this article we took a big picture look at how the Arcade Learning Environment is being used by the research community. We discussed the different evaluation methodologies that have been employed and how they have been frequently conflated in the literature. To further the progress in the field, we presented some methodological best practices and a new version of the Arcade Learning Environment that supports stochasticity and multiple game modes. We hope such methodological practices, with the new ALE, allow one to clearly distinguish between the different evaluation protocols. Also, we provide benchmark results following these methodological best practices that may serve as a point of comparison for future work in the ALE. We evaluated reinforcement learning algorithms that use linear and non-linear function approximation, and we hope to have promoted the discussion about sample efficiency by reporting algorithms' performance at different moments of the learning period. In the final part of this paper we concluded the big picture look we took by revisiting the challenges posed in the ALE's original article. We summarized the current state-of-the-art and we highlighted five problems we consider to remain open: representation learning, planning and model-learning, exploration, transfer learning, and off-policy learning.

\section*{Acknowledgements}
\label{ack}

The authors would like to thank David Silver and Tom Schaul for their thorough feedback on an earlier draft, and R\'emi Munos, Will Dabney, Mohammad Azar, Hector Geffner, Jean Harb, and Pierre-Luc Bacon for useful discussions. We would also like to thank the several contributors to the Arcade Learning Environment GitHub repository, specially Nicolas Carion for implementing the mode and difficult selection and Ben Goodrich for providing a Python interface to the ALE. Yitao Liang implemented, with Marlos C. Machado, the Blob-PROST features. This work was supported by grants from Alberta Innovates -- Technology Futures~(AITF), through the Alberta Machine Intelligence Institute (Amii), and by the NSF grant IIS-1552533. Computing resources were provided by Compute Canada through CalculQu\'ebec. 

\vskip 0.2in
\bibliography{jair_ale}
\bibliographystyle{theapa}

\clearpage

\appendix

\section{The Brute}\label{sec:brute}

The Brute is an algorithm designed to exploit features of the original Arcade Learning Environment. Although developed independently by some of the authors, it shares many similarities with the trajectory tree method of \citet{kearns99approximate}. The Brute relies on the following observations: 
\begin{itemize}
    \item{The ALE is deterministic, episodic, and guarantees a unique starting state, and}
    \item{in most Atari 2600 games, \emph{purpose} matters more than individual actions, \emph{i.e.}, most Atari 2600 games have important high-level goals, but individual actions have little impact.} 
\end{itemize}
This algorithm is crude but leads to competitive performance in a number of games.

\subsection{Determinism and starting configurations}

\begin{figure*}
\center{
\includegraphics[width=3in]{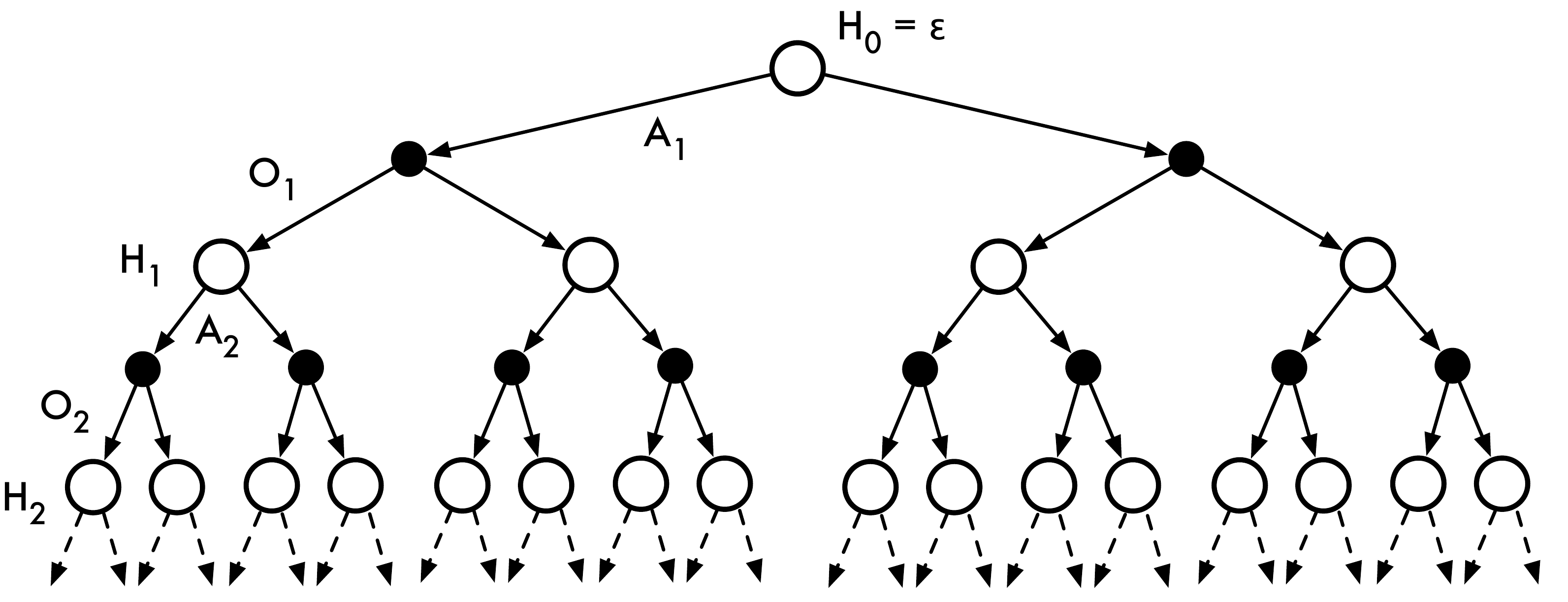}
}
\caption{History tree representation of an environment.\label{fig:history_tree}}
\end{figure*}

A history is a sequence of actions and observations $h_t = a_1, o_1, a_2, o_2, \dots, o_t$,
with the reward $r_t$ included in the observation $o_t$.\footnote{In the interest of legibility and symmetry, we follow here the convention of beginning histories with an action. Note, however, that the ALE provides the agent with an initial frame $o_0$.} 
Histories describe sequential interactions between an agent and its environment.
Although most of reinforcement learning focuses on a Markov state, a sufficient statistic
of the history, we may also reason directly about this history. This approach is particularly
convenient when the environment is partially observable \citep{kearns99approximate,evendar05reinforcement} or non-Markov \citep{hutter05universal}.
Given a history $h_t$, the transition function for an action $a$ and subsequent observation $o$ is 
\begin{equation*}
\Pr ( H_{t+1} = h_t, a, o \cbar H_t = h_t, A_t = a ) = \Pr ( O_{t+1} = o \cbar H_t = h_t, A_t = a ).
\end{equation*}
This transition function induces a Markov decision process over histories. This MDP is an infinite \emph{history tree} (Figure \ref{fig:history_tree}) whose
states correspond to distinct histories.

\begin{figure*}
\center{
\includegraphics[width=3in]{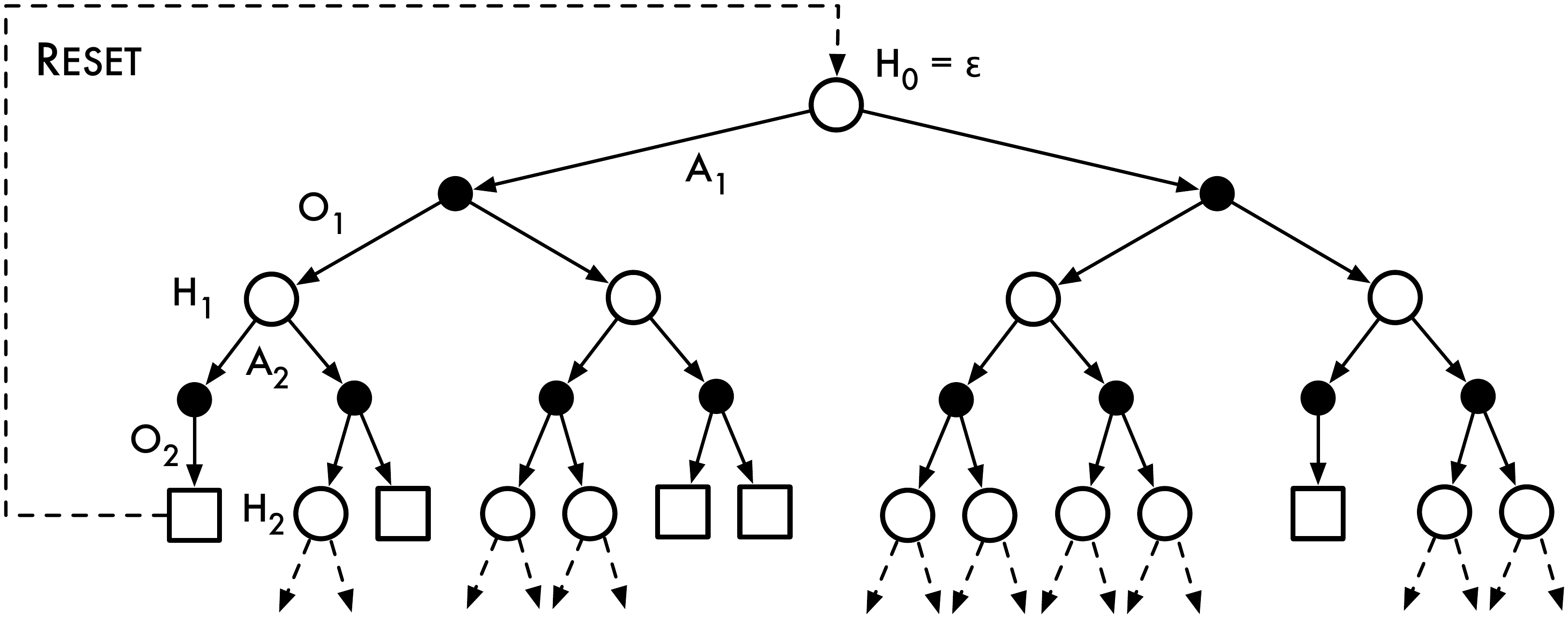}
}
\caption{In the episodic setting, the agent is reset after reaching a terminal state (represented by a square). We equate this reset with the empty history.\label{fig:history_tree_reset}}
\end{figure*}

An environment is deterministic if taking action $a$ from history $h$ always produces the same
observation. It is episodic when we have zero-valued, absorbing states called terminal states.
In the episodic setting learning proceeds by means of resets to one or many start states. Since
the agent is informed of this reset, we equate it with the empty history $\epsilon$~(Figure 
\ref{fig:history_tree_reset}). The Stella emulator is deterministic and, by the nature of Atari
2600 games, defines an episodic problem. 

Depending on the game, both software and hardware resets of the emulator may leave the system in a
number of initial configurations. These different configurations arise from changing timer values,
registers, and memory contents at reset. However, these effects are game-dependent and difficult to control. In fact, the ALE contains code to avoid these effects and guarantee a unique starting configuration. We will use the term \emph{reproducible} to describe an environment like the ALE that is deterministic, episodic, and has a unique starting configuration. 

Determinism simplifies the learning of an environment's transition model: a single sample from each state-action pair is sufficient. Reproducibility allows us to effectively perform experiments on the history tree, answering questions of the form ``what would happen if I performed this exact sequence of actions?'' Not unlike Monte-Carlo tree search in a deterministic domain, each experiment begins at the root of the history tree and selects actions until a terminal state is reached, observing rewards along the way. Although it is possible to do the same in any episodic environment, learning stochastic transitions and reward functions is harder, not only because they require more samples but also because the probability of reaching a particular state (\emph{i.e.}, a history) is exponentially small in its length.

\subsection{Value estimation in a history tree}

According to Bellman's optimality equation \cite{bellman57dynamic}, the optimal value of executing action $a$ in state $s$ is
\begin{equation*}
q_*(s, a) = \sum_{s'} p(s' \cbar s, a) \bigg(r(s,a, s') + \gamma \max_{b \in \cA} q_*(s', b)\bigg).
\end{equation*}
Given a full history tree of finite depth, estimating the value for any history-action pair is 
simply a matter of bottom-up dynamic programming, since all states (\emph{i.e.}, histories) are transient.
We can in fact leverage an important property of history trees:
Consider a partially known history tree for a deterministic environment and define $\hat q(h,a) = -\infty$ for any unknown history-action pair. Then the equation
\begin{equation}\label{eqn:history_tree_lower_bound_q}
\hat q(h,a) = \sum_{h'} p(h' \cbar h, a) \bigg(r(h,a,h') + \gamma \max_{b \in \cA} \hat q(h', b)\bigg)
\end{equation}
defines a lower bound on $q_*(h,a)$.

\begin{figure*}
\center{
\includegraphics[width=3in]{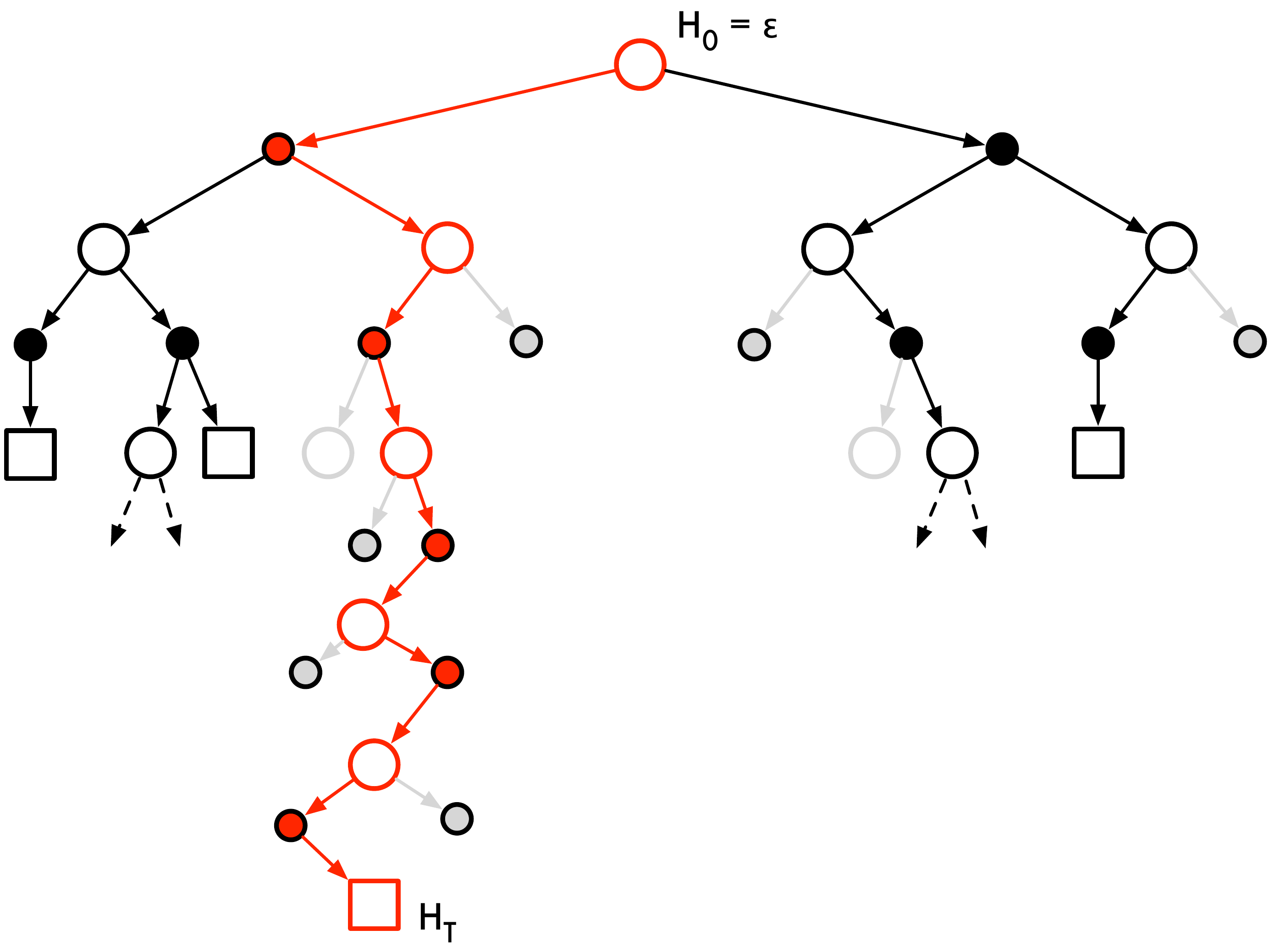}
}
\caption{A partially known history tree. Filled gray circles represent actions not yet taken by the agent, large gray circles unseen observations (when the environment is stochastic). The most recent episode is highlighted in red. The lower bound $\hat q(h,a)$ is updated by starting at $H_T$ and following the path back to the root.\label{fig:partial_history_tree_lower_bound}}
\end{figure*}

When learning proceeds in episodes, we can update the lower bound $\hat q(h,a)$ iteratively. We begin at the terminal node $h_T$ corresponding to the episode just played. We then follow the episode steps $a_{T-1}, h_{T-1}, a_{T-2}, h_{T-2}, \dots$ in reverse, updating $\hat q(h_t,a_t)$ along this path, up to and including the starting history-action pair $(\epsilon, a_1)$. Since no information has been gathered outside of this path, all other action-values must remain unchanged, and this procedure is correct. If $\pi(h) \doteq \argmax_{a \in \mathcal{A}} \hat q(h, a)$ is stored at each node, then updating one episode requires time $O(T)$. Figure \ref{fig:partial_history_tree_lower_bound} illustrates the inclusion of a new episode into a partial history tree.

The Brute maintains a partial history tree that contains all visited histories. Each node, associated
with a history, maintains an action-conditional transition function and reward function. Our
implementation maintains the maximum likelihood estimate for both functions. This allows us to
apply the Brute to stochastic environments, although $\hat q(h, a)$ is only guaranteed to be
a proper lower bound if the subtree rooted at $h$ is fully deterministic. This allowed us
to apply the exact same algorithm in the context of sticky actions~(Section~\ref{sec:determinism_stochasticity}). The value $\hat q(h, a)$ is maintained at each node and updated from
the maximum likelihood estimates at the end of each episode, as described above.

\subsection{Narrow exploration} 

In Atari 2600 games, most actions have little individual effect. An agent can thus be more
efficient if it focuses on a few narrow, promising trajectories rather than explore every detail
of its
environment. We may think of this focus as emphasizing purpose, \emph{i.e.}, achieving specific goals.
The sequence of actions which maximizes the lower bound $\hat q(h, a)$ at each node is one such 
purposeful path. Since exploration is less relevant at nodes which have been visited often, we
also progressively reduce the rate of exploration in the upper parts of the history tree.

To encourage the exploration of the most promising trajectory, the Brute's policy is an
$\epsilon$-greedy
policy over $\hat q(h,a)$: with probability $1 - \epsilon$, we choose one of the maximum-valued
actions (breaking ties uniformly at random), and with probability $\epsilon$ we select an action
uniformly at random. To encourage the exploration of narrow paths, $\epsilon$ is decreased
with the number of visits $n(h)$ to a particular node in the history tree. Specifically,
\begin{equation*}
\epsilon(h) = \min \left \{ \frac{0.05}{\log ( n(h) + 1 )}, 1.0 \right \} .
\end{equation*}

\section{Experimental Setup}~\label{sec:experimental_setup}

We used the same evaluation protocol, and parameters, in all experiments discussed in this article. In the next section we list the parameters used when defining the task in the Arcade Learning Environment. Later we discuss the parameters used by the Brute, Sarsa($\lambda$)~$+$~Blob-PROST, and DQN.

\subsection{Evaluation Protocol and Arcade Learning Environment Parameters}~\label{sec:eval_protocol}

We report our results aiming at evaluating the robustness of the learned policy and of the learning algorithm. All results we report for the Brute and for Sarsa($\lambda$) $+$ Blob-PROST are averaged over $24$ trials, and all results we report for DQN are averaged over $5$ trials. We evaluated DQN fewer times because its empirical validation is more expensive due to its requirement for specialized hardware (\emph{i.e.}, GPUs). We obtained the result of each trial by averaging over the last 100 episodes that led the agent to observe a total of $k$ frames. Along this article we reported results for $k$ equals to 10, 50, 100, and 200 million frames.

The unique parameter in the Arcade Learning Environment that is not fixed across all sections in this article is $\varsigma$, \emph{i.e.}, the amount of stochasticity present in the environment. We set $\varsigma$ to $0.0$ in Section~\ref{sec:eval_determinism} while we set $\varsigma$ to $0.25$ in the rest of the article. We do not use game-specific information. Episodes terminate after $5$ minutes of gameplay or when the agent has lost all of its lives. Agents have access to all $18$ primitive actions available in the ALE, not knowing if specific actions have any effect in the environment. Finally, all algorithms used a frame skip equals to $5$ when playing the games. We summarize all parameters that are shared across all methods in Table~\ref{tab:general_param}.

\begin{table}[t]
\centering
\caption{Parameters used to evaluate the Brute, DQN, and Sarsa($\lambda$) $+$ Blob-PROST.}
\footnotesize{
      \begin{tabular}{ p{3.1cm} | l | l }
            \textbf{Hyperparameter} &\textbf{Value} &\textbf{Description}\\
            \hline \hline
            
            \multirow{1}{3cm}{Action set}                      & \multirow{1}{2.2cm}{Full} & \multirow{1}{8.5cm}{$18$ actions are always available to the agent.}\\ \cdashline{1-3}[0.5pt/2pt]
            
            \multirow{1}{3cm}{Max. episode length}             & \multirow{1}{2.2cm}{18,000} & \multirow{1}{8.5cm}{Each episode lasts, at most, $5$ minutes (18,000 frames).} \\ \cdashline{1-3}[0.5pt/2pt]
            
            \multirow{1}{3cm}{Frame skip}                      & \multirow{1}{2.2cm}{5} & \multirow{1}{8.5cm}{Each action lasts $5$ time steps. See Section~\ref{sec:background} for details.} \\ \cdashline{1-3}[0.5pt/2pt]
            
            \multirow{2}{3cm}{Stochasticity ($\varsigma$)}     & \multirow{2}{2.2cm}{0.0 or 0.25} & \multirow{2}{8.5cm}{We used $\varsigma=0.25$ for all experiments, except for those in Section~\ref{sec:eval_determinism}, in which we used $\varsigma = 0.0$.} \\
            & & \\ \cdashline{1-3}[0.5pt/2pt]
            
            \multirow{2}{3cm}{Lives signal used}                & \multirow{2}{2.2cm}{False} & \multirow{2}{8.5cm}{We did not use the game-specific information about the number of lives the agent has at each time step.} \\
            & & \\ \cdashline{1-3}[0.5pt/2pt]
            
            \multirow{2}{3cm}{Number of episodes used for evaluation}     & \multirow{2}{2.2cm}{100} & \multirow{2}{8.5cm}{In the continual learning setting, we report the average score obtained in the last 100 episodes used for learning.} \\
            & & \\ \cdashline{1-3}[0.5pt/2pt]
            
            \multirow{2}{3cm}{Number of frames used for learning} & \multirow{2}{2.2cm}{10, 50, 100 and 200 million} & \multirow{2}{8.5cm}{We report scores obtained after each one of these four milestones.} \\
            & & \\ \cdashline{1-3}[0.5pt/2pt]
            
            \multirow{2}{3cm}{Number of trials ran}                & \multirow{2}{2.2cm}{24 or 5} & \multirow{2}{8.5cm}{The Brute and Sarsa($\lambda$) $+$ Blob-PROST were evaluated in 24 trials, DQN was evaluated in 5 trials.} \\
            & & \\
      \end{tabular}
}
\label{tab:general_param}
\end{table}

\subsection{Parameters used by the Brute}

The Brute has only two parameters to be set: $\gamma$ and $\epsilon$. We defined $\gamma = 1.0$ and $\epsilon = 0.005 / \log({n_i} + 2)$, where $n_i$ denotes the number of times we have seen the history $h_i$ (see Appendix~\ref{sec:brute} for details). An important implementation detail is that we used Spooky Hash\footnote{\url{http://burtleburtle.net/bob/hash/spooky.html}} as our hashing function. We do not average current and previous ALE screens as other methods do.

\subsection{Parameters used by DQN}

DQN was ran using the same parameters used in its original paper \cite{Mnih_Nature15}, with the exception of the frame skip, which we set to $5$ after preliminary experiments, and $\epsilon$, which we set to $0.01$ due to the absence of an evaluation phase. Also, we did not use game-specific information and we evaluated DQN in the continual learning setting, as discussed in Section~\ref{sec:eval_protocol}. Table~\ref{tab:dqn_param} lists the values of all DQN parameters used throughout this article.

\begin{table}[t]
\centering
\caption{Parameters used in the experiments evaluating DQN~\cite{Mnih_Nature15}. See reference for more details about the parameters listed below.}
\footnotesize{
      \begin{tabular}{ p{3.1cm} | l | l }
            \textbf{Hyperparameter} &\textbf{Value} &\textbf{Description}\\
            \hline \hline
            \multirow{1}{3cm}{Step-size ($\alpha$)}                       & \multirow{1}{2.2cm}{0.00025} & \multirow{1}{8.5cm}{Step-size used by RMSProp.} \\ \cdashline{1-3}[0.5pt/2pt]

            \multirow{1}{3cm}{Gradient momentum}                          & \multirow{1}{2.2cm}{0.95} & \multirow{1}{8.5cm}{Gradient momentum used by RMSProp.} \\ \cdashline{1-3}[0.5pt/2pt]

            \multirow{2}{3cm}{Squared gradient momentum}                  & \multirow{2}{2.2cm}{0.95} & \multirow{2}{8.5cm}{Squared gradient (denominator) momentum used by RMSProp.} \\
            & & \\ \cdashline{1-3}[0.5pt/2pt]

            \multirow{2}{3cm}{Min squared gradient}                       & \multirow{2}{2.2cm}{0.01} & \multirow{2}{8.5cm}{Constant added to the denominator of the RMSProp update.} \\
            & & \\ \cdashline{1-3}[0.5pt/2pt]

            \multirow{2}{3cm}{Discount factor ($\gamma$)}                 & \multirow{2}{2.2cm}{0.99} & \multirow{2}{8.5cm}{Discount factor used in Q-Learning update rule. Rewards are discounted by how far they are in time.} \\
            & & \\ \cdashline{1-3}[0.5pt/2pt]

            \multirow{1}{3cm}{Initial expl. rate ($\epsilon$)}            & \multirow{1}{2.2cm}{1.0} & \multirow{1}{8.5cm}{Probability a random action will be taken at each time step.} \\\cdashline{1-3}[0.5pt/2pt]

            \multirow{1}{3cm}{Final expl. rate ($\epsilon$)}              & \multirow{1}{2.2cm}{0.01} & \multirow{1}{8.5cm}{Probability a random action will be taken at each time step.} \\\cdashline{1-3}[0.5pt/2pt]

            \multirow{1}{3cm}{Final expl. frame}                          & \multirow{1}{2.2cm}{1,000,000} & \multirow{1}{8.5cm}{Number of frames over which $\epsilon$ is linearly annealead.} \\\cdashline{1-3}[0.5pt/2pt]

            \multirow{1}{3cm}{Minibatch size}                             & \multirow{1}{2.2cm}{32} & \multirow{1}{8.5cm}{Number of samples over which each update is computed.} \\\cdashline{1-3}[0.5pt/2pt]

            \multirow{2}{3cm}{Replay memory}                              & \multirow{2}{2.2cm}{1,000,000} & \multirow{2}{8.5cm}{The samples used in the algorithm's updates are drawn from the last 1 million recent frames.}\\
            & & \\ \cdashline{1-3}[0.5pt/2pt]

            \multirow{2}{3cm}{Replay start size}                          & \multirow{2}{2.2cm}{50,000} & \multirow{2}{8.5cm}{Number of frames over which a random policy is executed to first populate the replay memory.} \\
            & & \\ \cdashline{1-3}[0.5pt/2pt]

            \multirow{2}{3cm}{Agent history length}                       & \multirow{2}{2.2cm}{4} & \multirow{2}{8.5cm}{Number of most recent frames the agent observed that are given as input to the network.}  \\
            & & \\ \cdashline{1-3}[0.5pt/2pt]

            \multirow{2}{3cm}{Update frequency of target network }        & \multirow{2}{2.2cm}{4} & \multirow{2}{8.5cm}{Number of actions the agent selects between successive updates.} \\
            & & \\ \cdashline{1-3}[0.5pt/2pt]

            \multirow{2}{3cm}{Frame pooling}                              & \multirow{2}{2.2cm}{True} & \multirow{2}{8.5cm}{The observation received consists of the maximum pixel value between the previous and the current frame.} \\
            & & \\ \cdashline{1-3}[0.5pt/2pt]

            \multirow{2}{3cm}{Number of different colors}                 & \multirow{2}{2.2cm}{8} & \multirow{2}{8.5cm}{NTSC is the color palette in which each screen is encoded but at the end only the luminance channel is used.} \\
            & & \\
      \end{tabular}
}
\label{tab:dqn_param}
\end{table}

\subsection{Parameters used by Sarsa($\lambda$) $+$ Blob-PROST}

We evaluated Sarsa($\lambda$) $+$ Blob-PROST using $\alpha = 0.5$, $\lambda = 0.9$, and $\gamma = 0.99$. Agents followed an $\epsilon$-greedy policy ($\epsilon = 0.01$). We did not sweep most of the parameters, using the parameters reported by \citeauthor{Liang_AAMAS16}~\citeyear{Liang_AAMAS16}. However, we did verify, in preliminary experiments, the impact different values of frame skip have in this algorithm. We also verified whether color averaging impacts agents' performance. We decide to use a frame skip of $5$ and to average colors. For most games, averaging screen colors significantly improves the results, while the impact of different number of frames to skip varies across games. Table~\ref{tab:sarsa_param} summarizes, for Sarsa($\lambda$) $+$ Blob-PROST, all the parameters we use throughout this article.

\begin{table}[t]
\centering
\caption{Parameters used in the experiments evaluating Sarsa($\lambda$) $+$ Blob-PROST~\cite{Liang_AAMAS16}. See reference for more details about the parameters listed below.}
\footnotesize{
      \begin{tabular}{ p{3.1cm} | l | l }
            \textbf{Hyperparameter} &\textbf{Value} &\textbf{Description}\\
            \hline \hline
            \multirow{4}{3cm}{Step-size ($\alpha$)}                       & \multirow{4}{2.2cm}{0.50} & \multirow{4}{8.5cm}{Step-size used in Sarsa($\lambda$) update rule. At every time step we divide $\alpha$ by the largest number of active features we have seen so far. This reduces the step-size, avoiding divergence, while ensuring the step-size will never increase.} \\
            & & \\  
            & & \\ 
            & & \\ \cdashline{1-3}[0.5pt/2pt]
            
            \multirow{2}{3cm}{Discount factor ($\gamma$)}                 & \multirow{2}{2.2cm}{0.99} & \multirow{2}{8.5cm}{Discount factor used in Sarsa($\lambda$) update rule. Rewards are discounted by how far they are in time.} \\
            & & \\ \cdashline{1-3}[0.5pt/2pt]
            
            \multirow{1}{3cm}{Exploration rate ($\epsilon$)}              & \multirow{1}{2.2cm}{0.01} & \multirow{1}{8.5cm}{Probability a random action will be taken at each time step.} \\\cdashline{1-3}[0.5pt/2pt]
            
            \multirow{2}{3cm}{Eligibility traces decay rate ($\lambda$)}  & \multirow{2}{2.2cm}{0.90} & \multirow{2}{8.5cm}{Used in Sarsa($\lambda$) update rule. Encodes the trade-off between bias and variance.} \\
            & & \\ \cdashline{1-3}[0.5pt/2pt]
            
            \multirow{2}{3cm}{Eligibility threshold}                      & \multirow{2}{2.2cm}{0.01} &  \multirow{2}{8.5cm}{We set to $0$ any value in the eligibility trace vector that becomes smaller than this threshold.}\\
            & & \\ \cdashline{1-3}[0.5pt/2pt]
            
            \multirow{3}{3cm}{Feature set}                                & \multirow{3}{2.2cm}{Blob-PROST} & \multirow{3}{8.5cm}{Originally introduced by \citeauthor{Liang_AAMAS16}~\citeyear{Liang_AAMAS16}, Blob-PROST stands for Blob Pairwise Relative Offsets in Space and Time.} \\
            & & \\  
            & & \\ \cdashline{1-3}[0.5pt/2pt]
            
            \multirow{2}{3cm}{Color Averaging}                            & \multirow{2}{2.2cm}{True} & \multirow{2}{8.5cm}{The observation received is the average between the previous and the current frame.} \\
            & & \\ \cdashline{1-3}[0.5pt/2pt]
            
            \multirow{2}{3cm}{Grid width}                                 & \multirow{2}{2.2cm}{4} & \multirow{2}{8.5cm}{Each row of the game screen is divided into 40 tiles that are 4 pixels wide each.} \\
            & & \\ \cdashline{1-3}[0.5pt/2pt]
            
            \multirow{2}{3cm}{Grid height}                                & \multirow{2}{2.2cm}{7} & \multirow{2}{8.5cm}{Each column of the game screen is divided into 30 tiles that are 7 pixels high each.} \\
            & & \\ \cdashline{1-3}[0.5pt/2pt]
            
            \multirow{1}{3cm}{Neighborhood size}                         & \multirow{1}{2.2cm}{6} & \multirow{1}{8.5cm}{Tolerance used to detect blobs.} \\ \cdashline{1-3}[0.5pt/2pt]
            
            \multirow{2}{3cm}{Number of different colors}                 & \multirow{2}{2.2cm}{128} & \multirow{2}{8.5cm}{NTSC is the color palette in which each screen is encoded.} \\
            & & \\
      \end{tabular}
}
\label{tab:sarsa_param}
\end{table}

\section{Complete Benchmark Results}~\label{sec:full_benchmark_results}

We extend the results presented in Section~\ref{sec:benchmark} (Tables~\ref{tab:summary_blob_prost} and~\ref{tab:summary_dqn}) by reporting algorithms' performance in 60 games supported by the ALE. We used the evaluation protocol described in Appendix~\ref{sec:experimental_setup} when generating the results below. Table~\ref{tab:benchmark_blob_prost} summarizes the performance of Sarsa($\lambda$) $+$ Blob-PROST and Table~\ref{tab:benchmark_dqn} summarizes DQN's performance. The games originally used as training games by each method are highlighted with the $^\dagger$ symbol. In Table~\ref{tab:benchmark_blob_prost}, the list of games we used for training Sarsa($\lambda$) $+$ Blob-PROST is longer than the one in Table~\ref{tab:benchmark_dqn} because we are reporting the training games used by \citet{Liang_AAMAS16}, which was the setting we initially replicated. The columns with an asterisk will be filled in later.

\begin{table}[h]
\centering
\caption{Sarsa($\lambda$) $+$ Blob-PROST results across 60 games. See Appendix~\ref{sec:experimental_setup} for details.}
\scriptsize{
  \begin{tabular}{ l | r l ;{0.5pt/2pt} r l ;{0.5pt/2pt} r l ;{0.5pt/2pt} r l }
    Game &\multicolumn{2}{| c ;{0.5pt/2pt}}{10M frames} &\multicolumn{2}{;{0.5pt/2pt} c ;{0.5pt/2pt}}{50M frames} &\multicolumn{2}{;{0.5pt/2pt} c ;{0.5pt/2pt}}{100M frames} &\multicolumn{2}{;{0.5pt/2pt} c }{200M frames}\\
    \hline \hline
    \textsc{Alien}               &1,910.2   &(557.4)   &3,255.3   &(562.8)    &3,753.5   &(712.0)   &4,272.7   &(773.2)   \\ \hdashline[0.5pt/2pt]
    \textsc{Amidar}              &210.4     &(42.6)    &332.3     &(64.6)     &414.6     &(84.2)    &411.4     &(177.4)   \\ \hdashline[0.5pt/2pt]
    \textsc{Assault}             &435.9     &(94.8)    &651.9     &(148.7)    &851.7     &(185.4)   &1,049.4   &(182.7)   \\ \hdashline[0.5pt/2pt]
    \textsc{Asterix}$^\dagger$   &2,146.8   &(364.8)   &3,417.8   &(445.3)    &3,767.8   &(354.9)   &4,358.0   &(431.6)   \\ \hdashline[0.5pt/2pt]
    \textsc{Asteroids}           &1,350.1   &(259.5)   &1,378.1   &(233.0)    &1,443.4   &(218.1)   &1,524.1   &(191.2)   \\ \hdashline[0.5pt/2pt]
    \textsc{Atlantis}            &39,731.2  &(8,187.9) &41,833.5  &(23,356.0) &36,289.2  &(8,868.5) &38,057.5  &(8,455.2) \\ \hdashline[0.5pt/2pt]
    \textsc{Bank Heist}          &256.2     &(66.6)    &357.6     &(72.1)     &394.8     &(64.8)    &419.7     &(60.5)    \\ \hdashline[0.5pt/2pt]
    \textsc{Battle Zone}         &11,009.2  &(4,417.2) &19,178.3  &(3,293.4)  &22,419.2  &(4,204.4) &25,089.6  &(4,845.9) \\ \hdashline[0.5pt/2pt]
    \textsc{Beam Rider}$^\dagger$&1,200.2   &(242.9)   &1,859.2   &(391.9)    &2,126.1   &(523.7)   &2,234.0   &(471.5)   \\ \hdashline[0.5pt/2pt]
    \textsc{Berzerk}             &473.5     &(82.1)    &542.5     &(84.4)     &572.1     &(70.2)    &622.3     &(70.1)    \\ \hdashline[0.5pt/2pt]
    \textsc{Bowling}             &62.2      &(5.7)     &61.7      &(3.5)      &62.9      &(3.5)     &64.4      &(4.3)     \\ \hdashline[0.5pt/2pt]
    \textsc{Boxing}              &34.8      &(13.4)    &70.1      &(15.2)     &79.1      &(9.7)     &78.1      &(16.5)    \\ \hdashline[0.5pt/2pt]
    \textsc{Breakout}$^\dagger$  &12.3      &(1.5)     &16.8      &(1.5)      &18.6      &(1.7)     &20.2      &(1.9)     \\ \hdashline[0.5pt/2pt]
    \textsc{Carnival}            &2,206.2   &(855.7)   &4,207.5   &(857.7)    &4,959.7   &(935.9)   &3,489.8   &(2,621.5) \\ \hdashline[0.5pt/2pt]
    \textsc{Centipede}           &8,226.7   &(950.4)   &12,968.2  &(1,492.9)  &15,599.6  &(1,341.1) &1,189.3   &(1,040.4) \\ \hdashline[0.5pt/2pt]
    \textsc{Chopper Command}     &1,647.5   &(389.2)   &2,080.9   &(562.3)    &2,319.8   &(725.7)   &2,402.8   &(806.5)   \\ \hdashline[0.5pt/2pt]
    \textsc{Crazy Climber}       &32,518.8  &(3,868.1) &49,041.2  &(5,015.8)  &55,184.2  &(5,559.2) &60,471.0  &(5,534.9) \\ \hdashline[0.5pt/2pt]
    \textsc{Defender}            &5,775.3   &(890.9)   &7,343.6   &(1,607.2)  &8,863.3   &(1,380.2) &10,778.6  &(1,509.0) \\ \hdashline[0.5pt/2pt]
    \textsc{Demon Attack}        &385.4     &(144.8)   &628.9     &(96.9)     &921.5     &(91.5)    &1,272.2   &(253.6)   \\ \hdashline[0.5pt/2pt]
    \textsc{Double Dunk}         &-10.6     &(1.4)     &-8.5      &(0.8)      &-7.6      &(0.6)     &-6.9      &(0.5)     \\ \hdashline[0.5pt/2pt]
    \textsc{Elevator Action}     &3,228.9   &(4,415.4) &8,797.3   &(5,832.1)  &9,981.8   &(5,310.2) &11,147.8  &(4,291.3) \\ \hdashline[0.5pt/2pt]
    \textsc{Enduro}$^\dagger$    &120.3     &(49.8)    &241.3     &(28.6)     &275.1     &(13.0)    &294.0     &(8.0)     \\ \hdashline[0.5pt/2pt]
    \textsc{Fishing Derby}       &-87.4     &(4.9)     &-76.5     &(6.3)      &-73.2     &(6.7)     &-69.2     &(8.9)     \\ \hdashline[0.5pt/2pt]
    \textsc{Freeway}$^\dagger$   &29.9      &(1.6)     &31.8      &(0.3)      &31.9      &(0.3)     &31.8      &(0.3)     \\ \hdashline[0.5pt/2pt]
    \textsc{Frostbite}           &1,375.0   &(939.1)   &2,470.7   &(1,241.6)  &2,815.3   &(1,218.0) &3,207.2   &(1,040.4) \\ \hdashline[0.5pt/2pt]
    \textsc{Gopher}              &2,961.1   &(495.3)   &4,631.9   &(454.0)    &5,259.9   &(535.2)   &5,555.4   &(594.1)   \\ \hdashline[0.5pt/2pt]
    \textsc{Gravitar}            &629.8     &(201.5)   &863.5     &(255.6)    &979.5     &(340.1)   &1,150.0   &(397.5)   \\ \hdashline[0.5pt/2pt]
    \textsc{H.E.R.O.}            &9,452.6   &(2,433.1) &12,909.6  &(2,686.0)  &14,072.7  &(3,382.5) &14,910.2  &(3,887.6) \\ \hdashline[0.5pt/2pt]
    \textsc{Ice Hockey}          &-2.2      &(1.2)     &3.5       &(2.1)      &8.2       &(3.1)     &12.6      &(3.5)     \\ \hdashline[0.5pt/2pt]
    \textsc{James Bond}          &461.1     &(187.4)   &599.1     &(230.2)    &659.1     &(243.5)   &719.8     &(292.0)   \\ \hdashline[0.5pt/2pt]
    \textsc{Journey Escape}      &-5,569.6  &(1,231.0) &-5,115.5  &(5,821.8)  &-4,589.3  &(5,336.0) &*         &(*)       \\ \hdashline[0.5pt/2pt]
    \textsc{Kangaroo}            &1,305.6   &(555.7)   &2,442.5   &(1,282.4)  &3,152.8   &(1,546.3) &4,225.8   &(2,046.9) \\ \hdashline[0.5pt/2pt]
    \textsc{Krull}               &4,922.1   &(1,703.7) &6,762.2   &(5,168.9)  &7,491.5   &(5,823.9) &8,894.9   &(8,482.7) \\ \hdashline[0.5pt/2pt]
    \textsc{Kung-Fu Master}      &20,679.3  &(2,246.8) &23,548.3  &(2,926.8)  &26,745.6  &(3,281.5) &29,915.5  &(3,647.5) \\ \hdashline[0.5pt/2pt]
    \textsc{Montezuma's Revenge} &117.1     &(175.3)   &520.8     &(486.7)    &567.7     &(588.9)   &574.2     &(590.1)   \\ \hdashline[0.5pt/2pt]
    \textsc{Ms. Pac-Man}         &2,626.7   &(521.1)   &3,446.0   &(462.9)    &3,916.6   &(542.5)   &4,440.5   &(616.4)   \\ \hdashline[0.5pt/2pt]
    \textsc{Name This Game}      &4,626.2   &(284.3)   &6,164.2   &(357.0)    &6,219.8   &(1,821.2) &6,750.3   &(1,376.5) \\ \hdashline[0.5pt/2pt]
    \textsc{Phoenix}             &2,319.4   &(534.0)   &4,579.9   &(303.9)    &4,247.3   &(1,360.1) &5,197.0   &(374.5)   \\ \hdashline[0.5pt/2pt]
    \textsc{Pitfall!}            &-0.3      &(1.2)     &-0.1      &(0.3)      &0.0       &(0.0)     &0.0       &(0.0)     \\ \hdashline[0.5pt/2pt]
    \textsc{Pong}$^\dagger$      &1.8       &(3.9)     &10.9      &(3.3)      &12.6      &(2.8)     &14.5      &(2.0)     \\ \hdashline[0.5pt/2pt]
    \textsc{Pooyan}              &1,347.2   &(121.5)   &1,820.5   &(107.5)    &2,006.7   &(159.4)   &2,197.8   &(133.7)   \\ \hdashline[0.5pt/2pt]
    \textsc{Private Eye}         &36.7      &(46.3)    &36.2      &(49.2)     &27.9      &(44.9)    &44.2      &(49.3)    \\ \hdashline[0.5pt/2pt]
    \textsc{Q*bert}$^\dagger$    &3,535.9   &(745.2)   &4,605.7   &(567.3)    &5,931.9   &(1,174.4) &6,992.9   &(1,479.0) \\ \hdashline[0.5pt/2pt]
    \textsc{River Raid}          &4,141.9   &(574.4)   &7,399.5   &(492.5)    &8,988.3   &(1,154.3) &10,639.2  &(1,882.6) \\ \hdashline[0.5pt/2pt]
    \textsc{Road Runner}         &18,258.0  &(2,876.9) &23,380.2  &(5,940.3)  &28,453.4  &(3,227.4) &31,493.9  &(4,160.7) \\ \hdashline[0.5pt/2pt]
    \textsc{Robotank}            &21.3      &(3.4)     &25.0      &(2.8)      &26.3      &(2.7)     &27.3      &(2.3)     \\ \hdashline[0.5pt/2pt]
    \textsc{Seaquest}$^\dagger$  &788.2     &(225.2)   &1,201.6   &(178.4)    &1,319.2   &(356.5)   &1,402.9   &(328.3)   \\ \hdashline[0.5pt/2pt]
    \textsc{Skiing}              &-29,965.4 &(59.9)    &-29,955.6 &(50.7)     &-29,955.9 &(57.7)    &-29,940.2 &(102.2)   \\ \hdashline[0.5pt/2pt]
    \textsc{Solaris}             &480.9     &(185.1)   &585.5     &(213.3)    &704.2     &(264.3)   &807.3     &(216.2)   \\ \hdashline[0.5pt/2pt]
    \textsc{Space Invaders}$^\dagger$ &466.2 &(30.4)   &579.1     &(36.5)     &656.7     &(67.6)    &759.4     &(58.7)    \\ \hdashline[0.5pt/2pt]
    \textsc{Star Gunner}         &1,002.1   &(64.6)    &1,014.1   &(72.7)     &1,058.1   &(101.0)   &1,107.6   &(125.6)   \\ \hdashline[0.5pt/2pt]
    \textsc{Tennis}              &-0.1      &(0.1)     &-0.1      &(0.0)      &-0.1      &(0.0)     &-0.1      &(0.0)     \\ \hdashline[0.5pt/2pt]
    \textsc{Time Pilot}          &3,439.5   &(503.4)   &3,997.8   &(436.4)    &4,112.0   &(289.4)   &4,221.5   &(402.1)   \\ \hdashline[0.5pt/2pt]
    \textsc{Tutankham}           &122.2     &(3.5)     &152.7     &(16.0)     &88.9      &(64.1)    &91.5      &(63.3)    \\ \hdashline[0.5pt/2pt]
    \textsc{Up and Down}         &10,580.2  &(3,446.4) &10,049.1  &(9,340.7)  &11,514.5  &(11,988.8)&15,400.1  &(14,864.6)\\ \hdashline[0.5pt/2pt]
    \textsc{Venture}             &0.0       &(0.0)     &0.0       &(0.0)      &53.8      &(263.7)   &139.3     &(323.2)   \\ \hdashline[0.5pt/2pt]
    \textsc{Video Pinball}       &11,271.1  &(1,142.7) &13,259.3  &(1,327.2)  &14,334.7  &(1,097.4) &13,398.0  &(3,643.7) \\ \hdashline[0.5pt/2pt]
    \textsc{Wizard of Wor}       &1,975.9   &(471.4)   &2,738.8   &(613.3)    &3,247.5   &(713.0)   &2,043.5   &(801.3)   \\ \hdashline[0.5pt/2pt]
    \textsc{Yar's Revenge}       &4,961.2   &(1,200.2) &5,460.2   &(1,145.0)  &6,073.9   &(1,052.9) &7,257.8   &(1,884.8) \\ \hdashline[0.5pt/2pt]
    \textsc{Zaxxon}              &1,180.9   &(618.8)   &4,539.6   &(1,401.0)  &6,701.4   &(1,974.3) &8,166.8   &(3,979.8) \\
  \end{tabular}
}
\label{tab:benchmark_blob_prost}
\end{table}

\begin{table}[h]
\centering
\caption{DQN results across 60 games. See Appendix~\ref{sec:experimental_setup} for details.}
\scriptsize{
  \begin{tabular}{ l | r l ;{0.5pt/2pt} r l ;{0.5pt/2pt} r l ;{0.5pt/2pt} r l }
    Game &\multicolumn{2}{| c ;{0.5pt/2pt}}{10M frames} &\multicolumn{2}{;{0.5pt/2pt} c ;{0.5pt/2pt}}{50M frames} &\multicolumn{2}{;{0.5pt/2pt} c ;{0.5pt/2pt}}{100M frames} &\multicolumn{2}{;{0.5pt/2pt} c }{200M frames}\\
    \hline \hline
    \textsc{Alien}           &600.5   &(23.6)  &1,426.6  &(81.6)    &1,952.6   &(216.0)    &2,742.0   &(357.5)     \\ \hdashline[0.5pt/2pt]
    \textsc{Amidar}          &91.6    &(10.5)  &414.2    &(53.6)    &621.6     &(92.6)     &792.6     &(220.4)     \\ \hdashline[0.5pt/2pt]
    \textsc{Assault}         &688.9   &(16.0)  &1,327.5  &(83.9)    &1,433.9   &(126.6)    &1,424.6   &(106.8)     \\ \hdashline[0.5pt/2pt]
    \textsc{Asterix}$^\dagger$ &1,732.6 &(314.6) &3,122.6   &(96.4)    &3,423.4 &(213.6) &2,866.8 &(1,354.6) \\ \hdashline[0.5pt/2pt]
    \textsc{Asteroids}       &301.4   &(14.3)  &458.1    &(28.5)    &458.0     &(18.9)     &528.5     &(37.0)      \\ \hdashline[0.5pt/2pt]
    \textsc{Atlantis}        &6,639.4 &(208.4) &51,324.4 &(8,681.7) &291,134.7 &(31,575.2) &232,442.9 &(128,678.4) \\ \hdashline[0.5pt/2pt]
    \textsc{Bank Heist}      &32.3    &(6.5)   &448.2    &(104.8)   &740.7     &(130.6)    &760.0     &(82.3)      \\ \hdashline[0.5pt/2pt]
    \textsc{Battle Zone}     &2,428.3 &(200.4) &10,838.4 &(1,807.6) &15,048.5  &(2,372.0)  &20,547.5  &(1,843.0)   \\ \hdashline[0.5pt/2pt]
    \textsc{Beam Rider} $^\dagger$ &693.9 &(111.0) &4,551.5 &(849.1)   &4,977.2 &(292.2) &5,700.5 &(362.5)   \\ \hdashline[0.5pt/2pt]
    \textsc{Berzerk}         &434.5   &(51.2)  &457.5    &(9.4)     &470.0     &(24.5)     &487.2     &(29.9)      \\ \hdashline[0.5pt/2pt]
    \textsc{Bowling}         &28.7    &(0.8)   &29.4     &(1.8)     &32.8      &(3.6)      &33.6      &(2.7)       \\ \hdashline[0.5pt/2pt]
    \textsc{Boxing}          &18.6    &(3.8)   &71.7     &(2.7)     &77.9      &(0.5)      &72.7      &(4.9)       \\ \hdashline[0.5pt/2pt]
    \textsc{Breakout}        &14.2    &(1.2)   &75.1     &(4.3)     &57.9      &(14.6)     &35.1      &(22.6)      \\ \hdashline[0.5pt/2pt]
    \textsc{Carnival}        &588.5   &(47.0)  &2,131.6  &(534.3)   &4,621.9   &(191.0)    &4,803.8   &(189.0)     \\ \hdashline[0.5pt/2pt]
    \textsc{Centipede}       &3,075.2 &(381.1) &2,280.0  &(184.2)   &2,555.2   &(195.1)    &2,838.9   &(225.3)     \\ \hdashline[0.5pt/2pt]
    \textsc{Chopper Command} &841.4   &(144.3) &2,104.8  &(327.7)   &3,288.1   &(339.2)    &4,399.6   &(401.5)     \\ \hdashline[0.5pt/2pt]
    \textsc{Crazy Climber}   &43,716.6 &(2,571.2) &80,599.6 &(4,209.8) &64,807.3 &(26,100.0) &78,352.1 &(1,967.3)  \\ \hdashline[0.5pt/2pt]
    \textsc{Defender}        &2,409.9 &(78.6) &2,525.7   &(124.0)   &2,711.6   &(96.8)     &2,941.3   &(106.2)     \\ \hdashline[0.5pt/2pt]
    \textsc{Demon Attack}    &154.8   &(11.5) &3,744.6   &(688.9)   &4,556.5   &(947.2)    &5,182.0   &(778.0)     \\ \hdashline[0.5pt/2pt]
    \textsc{Double Dunk}     &-20.9   &(0.3)  &-18.4     &(1.2)     &-15.6     &(1.6)      &-8.7      &(4.5)       \\ \hdashline[0.5pt/2pt]
    \textsc{Elevator Action} &6.7     &(13.3) &4.5       &(9.0)     &4.7       &(9.4)      &6.0       &(10.4)      \\ \hdashline[0.5pt/2pt]
    \textsc{Enduro}          &473.2   &(22.3) &578.0     &(79.6)    &597.4     &(153.1)    &688.2     &(32.4)      \\ \hdashline[0.5pt/2pt]
    \textsc{Fishing Derby}   &-63.1   &(7.8)  &7.5       &(4.1)     &12.2      &(1.4)      &10.2      &(1.9)       \\ \hdashline[0.5pt/2pt]
    \textsc{Freeway}$^\dagger$ &13.8  &(8.1)     &31.7      &(0.7)     &32.4    &(0.3)   &33.0    &(0.3)     \\ \hdashline[0.5pt/2pt]
    \textsc{Frostbite}       &241.8   &(30.8) &292.5     &(28.8)    &274.3     &(8.8)      &279.6     &(13.9)      \\ \hdashline[0.5pt/2pt]
    \textsc{Gopher}          &679.6   &(35.2) &2,233.7   &(123.1)   &2,988.8   &(514.4)    &3,925.5   &(521.4)     \\ \hdashline[0.5pt/2pt]
    \textsc{Gravitar}        &79.5    &(8.0)  &109.3     &(3.1)     &118.5     &(22.0)     &154.9     &(17.7)      \\ \hdashline[0.5pt/2pt]
    \textsc{H.E.R.O.}        &1,667.9 &(1,107.8) &11,564.0 &(3,722.4) &14,684.7 &(1,840.6) &18,843.3  &(2,234.9)   \\ \hdashline[0.5pt/2pt]
    \textsc{Ice Hockey}      &-15.1   &(0.3)   &-8.9     &(1.7)     &-4.4      &(2.0)      &-3.8      &(4.7)       \\ \hdashline[0.5pt/2pt]
    \textsc{James Bond}      &30.7    &(6.0)   &191.4    &(144.9)   &517.2     &(35.8)     &581.0     &(21.3)      \\ \hdashline[0.5pt/2pt]
    \textsc{Journey Escape}  &-2,220.0 &(176.1) &-2,409.7 &(341.2)  &-2,959.0  &(383.9)    &-3,503.0  &(488.5)     \\ \hdashline[0.5pt/2pt]
    \textsc{Kangaroo}        &298.6   &(56.1)  &8,878.8  &(2,886.1) &12,846.9  &(688.3)    &12,291.7  &(1,115.9)   \\ \hdashline[0.5pt/2pt]
    \textsc{Krull}           &4,424.7 &(492.7) &6,035.6  &(248.6)   &6,589.8   &(264.4)    &6,416.0   &(128.5)     \\ \hdashline[0.5pt/2pt]
    \textsc{Kung-Fu Master}  &9,468.1 &(1,975.9) &17,537.4 &(1,128.8) &17,772.3 &(3,423.3) &16,472.7  &(2,892.7)   \\ \hdashline[0.5pt/2pt]
    \textsc{Montezuma's Revenge} &0.2 &(0.4)     &0.2      &(0.4)     &0.0      &(0.0)     &0.0       &(0.0)       \\ \hdashline[0.5pt/2pt]
    \textsc{Ms. Pac-Man}     &1,675.5 &(41.9)    &2,626.1  &(139.8)   &2,964.9  &(100.8)   &3,116.2   &(141.2)     \\ \hdashline[0.5pt/2pt]
    \textsc{Name This Game}  &2,265.6 &(171.0)   &4,105.4  &(932.3)   &4,105.6  &(653.5)   &3,925.2   &(660.2)     \\ \hdashline[0.5pt/2pt]
    \textsc{Phoenix}         &1,501.2 &(278.1)   &3,174.0  &(543.5)   &2,607.1  &(644.1)   &2,831.0   &(581.0)     \\ \hdashline[0.5pt/2pt]
    \textsc{Pitfall!}        &-24.9   &(14.8)    &-28.2    &(13.0)    &-23.3    &(9.6)     &-21.4     &(3.2)       \\ \hdashline[0.5pt/2pt]
    \textsc{Pong}            &-15.9   &(1.0)     &12.2     &(1.0)     &15.2     &(0.7)     &15.1      &(1.0)       \\ \hdashline[0.5pt/2pt]
    \textsc{Pooyan}          &2,278.9 &(273.7)   &3,528.9  &(256.3)   &3,387.8  &(182.8)   &3,700.4   &(349.5)     \\ \hdashline[0.5pt/2pt]
    \textsc{Private Eye}     &81.6    &(15.6)    &60.4     &(92.4)    &1,447.4  &(2,567.9) &3,967.5   &(5,540.6)   \\ \hdashline[0.5pt/2pt]
    \textsc{Q*bert}          &674.7   &(53.6)    &3,142.1  &(1,238.7) &7,585.4  &(2,787.4) &9,875.5   &(1,385.3)   \\ \hdashline[0.5pt/2pt]
    \textsc{River Raid}      &3,166.2 &(125.2)   &8,738.1  &(500.0)   &10,733.1 &(229.9)   &10,210.4  &(435.0)     \\ \hdashline[0.5pt/2pt]
    \textsc{Road Runner}     &14,742.2 &(1,553.4) &37,271.7 &(1,234.5) &41,918.4 &(1,762.5) &42,028.3 &(1,492.0)   \\ \hdashline[0.5pt/2pt]
    \textsc{Robotank}        &4.1     &(0.3)     &28.4     &(1.4)     &38.0     &(1.6)     &58.0      &(6.4)       \\ \hdashline[0.5pt/2pt]
    \textsc{Seaquest}$^\dagger$ &311.5 &(36.9)   &1,430.8   &(162.3)   &1,573.4 &(561.4) &1,485.7 &(740.8)   \\ \hdashline[0.5pt/2pt]
    \textsc{Skiing}          &-20,837.5 &(1,550.2) &-17,545.5 &(4,041.5) &-13,365.1 &(800.7) &-12,446.6 &(1,257.9) \\ \hdashline[0.5pt/2pt]
    \textsc{Solaris}         &1,030.2  &(40.3)   &977.7    &(112.5)   &783.4    &(55.3)    &1,210.0   &(148.3)     \\ \hdashline[0.5pt/2pt]
    \textsc{Space Invaders} $^\dagger$ &211.6  &(14.8)    &686.6     &(37.0)    &787.2   &(173.3) &823.6   &(335.0)   \\ \hdashline[0.5pt/2pt]
    \textsc{Star Gunner}     &603.0    &(28.0)   &1,492.3  &(79.7)    &11,590.5 &(4,658.9) &39,269.9  &(5,298.8)   \\ \hdashline[0.5pt/2pt]
    \textsc{Tennis}          &-23.8    &(0.1)    &-23.9    &(0.1)     &-23.9    &(0.0)     &-23.9     &(0.0)       \\ \hdashline[0.5pt/2pt]
    \textsc{Time Pilot}      &1,078.8  &(60.3)   &1,068.1  &(138.8)   &1,330.7  &(177.1)   &2,061.8   &(228.8)     \\ \hdashline[0.5pt/2pt]
    \textsc{Tutankham}       &56.5     &(10.0)   &64.9     &(12.6)    &65.1     &(11.9)    &60.0      &(12.7)      \\ \hdashline[0.5pt/2pt]
    \textsc{Up and Down}     &4,378.4  &(172.5)  &6,718.3  &(671.2)   &5,962.8  &(618.7)   &4,750.7   &(1,007.5)   \\ \hdashline[0.5pt/2pt]
    \textsc{Venture}         &24.4     &(46.9)   &21.4     &(15.1)    &4.4      &(5.4)     &3.2       &(4.7)       \\ \hdashline[0.5pt/2pt]
    \textsc{Video Pinball}   &4,009.3  &(271.9)  &7,817.0  &(1,884.4) &16,626.2 &(3,740.6) &15,398.5  &(2,126.1)   \\ \hdashline[0.5pt/2pt]
    \textsc{Wizard of Wor}   &184.2    &(22.0)   &1,377.4  &(71.0)    &1,440.6  &(237.3)   &2,231.1   &(820.8)     \\ \hdashline[0.5pt/2pt]
    \textsc{Yar's Revenge}   &7,261.4  &(777.1)  &10,344.8 &(452.4)   &10,312.3 &(528.9)   &13,073.4  &(1,961.8)   \\ \hdashline[0.5pt/2pt]
    \textsc{Zaxxon}          &53.5     &(51.0)   &672.3    &(748.5)   &1,638.2  &(784.0)   &3,852.1   &(1,120.7)   \\
  \end{tabular}
}
\label{tab:benchmark_dqn}
\end{table}

\clearpage

\section{Number of Game Modes and Difficulties in the Games Supported by the Arcade Learning Environment}~\label{sec:flavors}

\begin{table}[h]
\caption{Atari 2600 games supported by the Arcade Learning Environment and the respective number of modes and difficulties available in each game. Modes only playable by two-players have been excluded.}
\small{
\begin{tabular}{l| c;{0.5pt/2pt} c | l| c;{0.5pt/2pt} c }
\textsc{Game} &\# Modes & \# Diffic. &\textsc{Game} &\# Modes &\# Diffic.\\
\hline \hline
\textsc{Alien} &4 &4 &\textsc{Journey Escape} &1 &2\\ \hdashline[0.5pt/2pt]
\textsc{Amidar} &1 &2 &\textsc{Kangaroo} &2 &1\\ \hdashline[0.5pt/2pt]
\textsc{Assault} &1 &1 &\textsc{Krull} &4 &1\\ \hdashline[0.5pt/2pt]
\textsc{Asterix} &1 &1 &\textsc{Kung Fu Master} &1 &1\\ \hdashline[0.5pt/2pt]
\textsc{Asteroids} &33 &2 &\textsc{Montezuma Revenge} &1 &1\\ \hdashline[0.5pt/2pt]
\textsc{Atlantis} &4 &1 &\textsc{Ms. Pac-Man} &4 &1\\ \hdashline[0.5pt/2pt]
\textsc{Bank Heist} &8 &4 &\textsc{Name this Game} &3 &2\\ \hdashline[0.5pt/2pt]
\textsc{Battle Zone} &3 &1 &\textsc{Phoenix} &1 &1\\ \hdashline[0.5pt/2pt]
\textsc{Beam Rider} &1 &2 &\textsc{Pitfall} &1 &1\\ \hdashline[0.5pt/2pt]
\textsc{Berzerk} &12 &1 &\textsc{Pong} &2 &2\\ \hdashline[0.5pt/2pt]
\textsc{Bowling} &3 &2 &\textsc{Pooyan} &4 &1\\ \hdashline[0.5pt/2pt]
\textsc{Boxing} &1 &4 &\textsc{Private Eye} &5 &4\\ \hdashline[0.5pt/2pt]
\textsc{Breakout} &12 &2 &\textsc{Q*bert} &1 &2\\ \hdashline[0.5pt/2pt]
\textsc{Carnival} &1 &1 &\textsc{River Raid} &1 &2\\ \hdashline[0.5pt/2pt]
\textsc{Centipede} &2 &1 &\textsc{Road Runner} &1 &1\\ \hdashline[0.5pt/2pt]
\textsc{Chopper Command} &2 &2 &\textsc{Robot Tank} &1 &1\\ \hdashline[0.5pt/2pt]
\textsc{Crazy Climber} &4 &2 &\textsc{Seaquest} &1 &2\\ \hdashline[0.5pt/2pt]
\textsc{Defender} &10 &2 &\textsc{Skiing} &10 &1\\ \hdashline[0.5pt/2pt]
\textsc{Demon Attack} &4 &2 &\textsc{Solaris} &1 &1\\ \hdashline[0.5pt/2pt]
\textsc{Double Dunk} &16 &1 &\textsc{Space Invader} &16 &2\\ \hdashline[0.5pt/2pt]
\textsc{Elevator Action} &1 &1 &\textsc{Star Gunner} &4 &1\\ \hdashline[0.5pt/2pt]
\textsc{Enduro} &1 &1 &\textsc{Tennis} &2 &4\\ \hdashline[0.5pt/2pt]
\textsc{Fishing Derby} &1 &4 &\textsc{Time Pilot} &1 &3\\ \hdashline[0.5pt/2pt]
\textsc{Freeway} &8 &2 &\textsc{Tutankham} &4 &1\\ \hdashline[0.5pt/2pt]
\textsc{Frostbite} &2 &1 &\textsc{UpNDown} &1 &4\\ \hdashline[0.5pt/2pt]
\textsc{Gopher} &2 &2 &\textsc{Venture} &1 &4\\ \hdashline[0.5pt/2pt]
\textsc{Gravitar} &5 &1 &\textsc{Video Pinball} &2 &2\\ \hdashline[0.5pt/2pt]
\textsc{Hero} &5 &1 &\textsc{Wizard Of Wor} &1 &2\\ \hdashline[0.5pt/2pt]
\textsc{Ice Hockey} &2 &4 &\textsc{Yar's Revenge} &4 &2\\ \hdashline[0.5pt/2pt]
\textsc{James Bond} &2 &1 &\textsc{Zaxxon} &4 &1\\
\end{tabular}
}
\end{table}

\end{document}